\documentclass[journal]{IEEEtran}

\usepackage{graphicx}
\usepackage{hyperref}

\usepackage[dvipsnames]{xcolor}
\begin{document}

\title{Deep Convolutional Neural Networks for Computer-Aided Detection: CNN Architectures, Dataset Characteristics and Transfer Learning}

\author{Hoo-Chang Shin, {\em Member, IEEE}, Holger R. Roth, Mingchen Gao, Le Lu, {\em Senior Member, IEEE}, Ziyue Xu, \\ Isabella Nogues, Jianhua Yao,  Daniel Mollura, Ronald M. Summers*  
\thanks{Hoo-Chang Shin, Holger R. Roth, Le Lu, Isabella Nogues, Jianhua Yao and Ronald M. Summers are with the Imaging Biomarkers and Computer-Aided Diagnosis Laboratory; Mingchen Gao, Ziyue Xu and Daniel Mollura are with Center for Infectious Disease Imaging, Le Lu, Jianhua Yao and Ronald M. Summers are also with Clinical Image Processing Service, Radiology and Imaging Sciences Department, National Institutes of Health Clinical Center, Bethesda, MD 20892-1182, USA. {\it Holger R. Roth and Mingchen Gao contributed equally to this work. Asterisk indicates corresponding author. } e-mail: \{hoochang.shin, le.lu, rms\}@nih.gov. Copyright (c) 2010 IEEE. Personal use of this material is permitted. However, permission to use this material for any other purposes must be obtained from the IEEE by sending a request to pubs-permissions@ieee.org.}% <-this % stops a space
}

\maketitle

\begin{abstract}

Remarkable progress has been made in image recognition, primarily due to the availability of large-scale annotated datasets (i.e. ImageNet) and the revival of deep convolutional neural networks (CNN).
CNNs enable learning data-driven, highly representative, layered hierarchical image features from sufficient training data.
However, obtaining datasets as comprehensively annotated as ImageNet in the medical imaging domain remains a challenge.
There are currently three major techniques that successfully employ CNNs to medical image classification: training the CNN from scratch, using off-the-shelf pre-trained CNN features, and conducting unsupervised CNN pre-training with supervised fine-tuning.
Another effective method is transfer learning, i.e., fine-tuning CNN models (supervised) pre-trained from natural image dataset to medical image tasks (although domain transfer between two medical image datasets is also possible).

In this paper, we exploit three important, but previously understudied factors of employing deep convolutional neural networks to computer-aided detection problems.
We first explore and evaluate different CNN architectures. The studied models contain 5 thousand to 160 million parameters, and vary in numbers of layers.
We then evaluate the influence of dataset scale and spatial image context on performance.
Finally, we examine when and why transfer learning from pre-trained ImageNet (via fine-tuning) can be useful.
We study two specific computer-aided detection (CADe) problems, namely thoraco-abdominal lymph node (LN) detection and interstitial lung disease (ILD) classification. We achieve the state-of-the-art performance on the mediastinal LN detection, with 85\% sensitivity at 3 false positive per patient, and report the first five-fold cross-validation classification results on predicting axial CT slices with ILD categories. Our extensive empirical evaluation, CNN model analysis and valuable insights can be extended to the design of high performance CAD systems for other medical imaging tasks.

\end{abstract}


% \fi
\IEEEpeerreviewmaketitle

\section{Introduction}
Tremendous progress has been made in image recognition, primarily due to the availability of large-scale annotated datasets (i.e. ImageNet \cite{deng2009imagenet,Russakovsky2014ILSVRC}) and the recent revival of deep convolutional neural networks (CNN) \cite{LeCun1998gradient,krizhevsky2012imagenet}.
For data-driven learning, large-scale well-annotated datasets with representative data distribution characteristics are crucial to learning more accurate or generalizable models \cite{krizhevsky2009learning,krizhevsky2012imagenet}. 
Unlike previous image datasets used in computer vision, ImageNet \cite{deng2009imagenet} offers a very comprehensive database of more than 1.2 million categorized natural images of 1000+ classes.
The CNN models trained upon this database serve as the backbone for significantly improving many object detection and image segmentation problems using other datasets \cite{Girshick15,He2015SPPNet}, e.g., PASCAL \cite{Everingham2015Pascal} and medical image categorization \cite{Ginneken15,Bar2015,Shin2015Interleaved,Ciompi2015automatic}.
However, there exists no large-scale annotated medical image dataset comparable to ImageNet, as data acquisition is difficult, and quality annotation is costly.

There are currently three major techniques that successfully employ CNNs to medical image classification: \textit{1)} training the ``CNN from scratch'' \cite{menze2014multimodal,pan2015brain,shen2015multi,carneiro2015unregistered,wolterink2015automatic}; \textit{2)} using ``off-the-shelf CNN'' features (without retraining the CNN) as complementary information channels to existing hand-crafted image features, for Chest X-rays \cite{Bar2015} and CT lung nodule identification \cite{Ginneken15,Ciompi2015automatic}; and \textit{3)} performing unsupervised pre-training on natural or medical images and fine-tuning on medical target images using CNN or other types of deep learning models \cite{schlegl2014unsupervised,hofmanninger2015mapping,Carneiro2013,Shen2014}.
A decompositional 2.5D view resampling and an aggregation of random view classification scores are used to eliminate the ``curse-of-dimensionality'' issue in \cite{roth2015improving}, in order to acquire a sufficient number of training image samples.

Previous studies have analyzed three-dimensional patch creation for LN detection \cite{barbu2012automatic,feulner2013lymph}, atlas creation from chest CT \cite{feuerstein2012mediastinal} and the extraction of multi-level image features \cite{lu2014computer,Lu2011}.
At present, there are several extensions or variations of the decompositional view representation introduced in \cite{roth2015improving,Lu2008}, such as: using a novel vessel-aligned multi-planar image representation for pulmonary embolism detection \cite{tajbakhsh2015computer}, fusing unregistered multiview for mammogram analysis \cite{carneiro2015unregistered} and classifying pulmonary peri-fissural nodules via an ensemble of 2D views \cite{Ciompi2015automatic}.

Although natural images and medical images differ significantly, conventional image descriptors developed for object recognition in natural images, such as the scale-invariant feature transform (SIFT) \cite{lowe2004distinctive} and the histogram of oriented gradients (HOG) \cite{dalal2005histograms}, have been widely used for object detection and segmentation in medical image analysis. Recently, ImageNet pre-trained CNNs have been used for chest pathology identification and detection in X-ray and CT modalities \cite{Bar2015,Ginneken15,Ciompi2015automatic}.
They have yielded the best performance results by integrating low-level image features (e.g., GIST \cite{torralba2008small}, bag of visual words (BoVW) and bag-of-frequency \cite{Ciompi2015automatic}). However, the fine-tuning of an ImageNet pre-trained CNN model on medical image datasets has not yet been exploited.

In this paper, we exploit three important, but previously under-studied factors of employing deep convolutional neural networks to computer-aided detection problems.
Particularly, we explore and evaluate different CNN architectures varying in width (ranging from 5 thousand to 160 million parameters) and depth (various numbers of layers), describe the effects of varying dataset scale and spatial image context on performance, and discuss when and why transfer learning from pre-trained ImageNet CNN models can be valuable. We further verify our hypothesis by inheriting and adapting rich hierarchical image features \cite{krizhevsky2009learning,szegedy2014going} from the large-scale ImageNet dataset for computer aided diagnosis (CAD). We also explore CNN architectures of the most studied seven-layered ``AlexNet-CNN'' \cite{krizhevsky2012imagenet}, a shallower ``Cifar-CNN'' \cite{roth2015improving}, and a much deeper version of ``GoogLeNet-CNN'' \cite{szegedy2014going} (with our modifications on CNN structures). This study is partially motivated by recent studies \cite{Chatfield2014return,chatfield2011devil} in computer vision. The thorough quantitative analysis and evaluation on deep CNN \cite{Chatfield2014return} or sparsity image coding methods \cite{chatfield2011devil} elucidate the emerging techniques of the time and provide useful suggestions for their future stages of development, respectively.

Two specific computer-aided detection (CADe) problems, namely thoraco-abdominal lymph node (LN) detection and interstitial lung disease (ILD) classification are studied in this work. On mediastinal LN detection, we surpass all currently reported results.
We obtain $86\%$ sensitivity on 3 false positives (FP) per patient, versus the prior state-of-art sensitivities of $78\%$ \cite{seff20152d} (stacked shallow learning) and $70\%$ \cite{roth2015improving} (CNN), as prior state-of-the-art. For the first time, ILD classification results under the patient-level five-fold cross-validation protocol (CV5) are investigated and reported. The ILD dataset \cite{depeursinge2012building} contains 905 annotated image slices with 120 patients and 6 ILD labels. Such sparsely annotated datasets are generally difficult for CNN learning, due to the paucity of labeled instances.

Evaluation protocols and details are critical to deriving significant empirical findings \cite{Chatfield2014return}.
Our experimental results suggest that different CNN architectures and dataset re-sampling protocols are critical for the LN detection tasks where the amount of labeled training data is sufficient and spatial contexts are local.
Since LN images are more flexible than ILD images with respect to resampling and reformatting, LN datasets may be more readily augmented by such image transformations. As a result, LN datasets contain more training and testing data instances (due to data auugmentation) than ILD datasets. They nonetheless remain less  comprehensive than natural image datasets, such as ImageNet.
Fine-tuning ImageNet-trained models for ILD classification is clearly advantageous and yields early promising results, when the amount of labeled training data is highly insufficient and multi-class categorization is used, as opposed to the LN dataset's binary class categorization.
Another significant finding is that CNNs trained from scratch or fine-tuned from ImageNet models consistently outperform CNNs that merely use off-the-shelf CNN features, in both the LN and ILD classification problems. We further analyze, via CNN activation visualizations, when and why transfer learning from non-medical to medical images in CADe problems can be valuable.

\section{Datasets and Related Work}\label{sec:datasets}

We employ CNNs (with the characteristics defined above) to thoraco-abdominal lymph node (LN) detection (evaluated separately on the mediastinal and abdominal regions) and interstitial lung disease (ILD) detection. For LN detection, we use randomly sampled 2.5D views in CT \cite{roth2015improving}. We use 2D CT slices \cite{song2013feature,song2015large,gao2014holistic} for ILD detection.
We then evaluate and compare CNN performance results.

Until the detection aggregation approach \cite{roth2015improving,seff20142d}, thoracoabdominal lymph node (LN) detection via CADe mechanisms has yielded poor performance results. In \cite{roth2015improving}, each 3D LN candidate produces up to 100 random 2.5D orthogonally sampled images or views which are then used to train an effective CNN model. The best performance on abdominal LN detection is achieved at $83\%$ recall on 3FP per patient \cite{roth2015improving}, using a ``Cifar-10'' CNN. Using the thoracoabdominal LN detection datasets \cite{roth2015improving}, {\em we aim to surpass this CADe performance level, by testing different CNN architectures, exploring various dataset re-sampling protocols, and applying transfer learning from ImageNet pre-trained CNN models.}

Interstitial lung disease (ILD) comprises more than 150 lung diseases affecting the interstitium, which can severely impair the patient's ability to breathe. Gao et al. \cite{gao2014holistic} investigate the ILD classification problem in two scenarios: \textit{1)} slice-level classification: assigning a holistic two-dimensional axial CT slice image with its occurring ILD disease label(s); and \textit{2)} patch-level classification: \textit{a/} sampling patches within the 2D ROIs (Regions of Interest provided by \cite{depeursinge2012building}), then \textit{b/} classifying patches into seven category labels ( six disease labels and one ``healthy'' label). Song et al. \cite{song2013feature,song2015large} only address the second sub-task of patch-level classification under the ``leave-one-patient-out'' (LOO) criterion. By training on the moderate-to-small scale ILD dataset \cite{depeursinge2012building}, {\em our main objective is to exploit and benchmark CNN based ILD classification performances under the CV5 metric (which is more realistic and unbiased than LOO \cite{song2013feature,song2015large} and hard-split \cite{gao2014holistic}), with and without transfer learning.}

\begin{figure}[t]
\begin{center}
   %\vspace{.27cm}
   \includegraphics[width=1\linewidth]{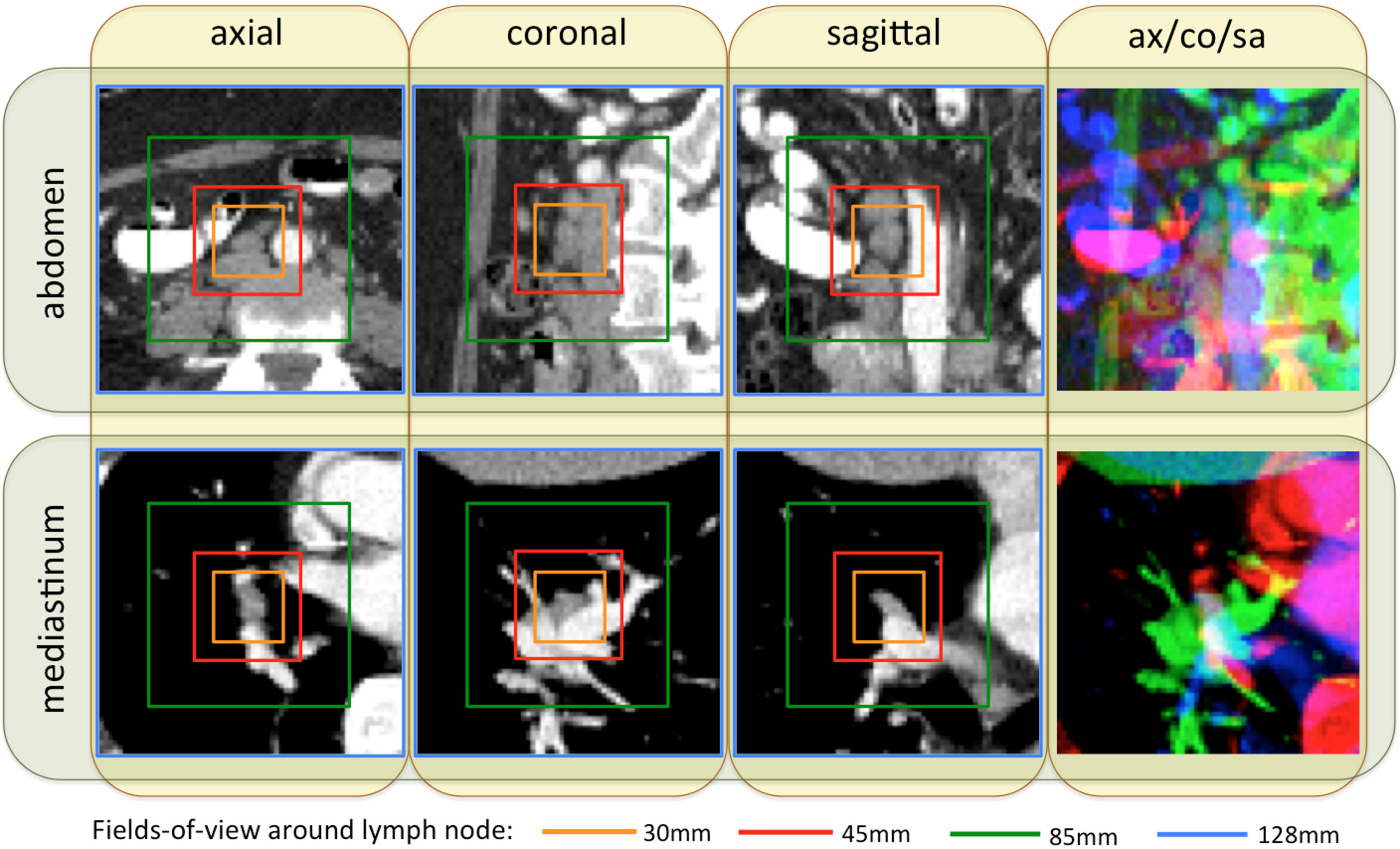}
\end{center}
   \caption{Some examples of abdominal and mediastinal lymph nodes sampled on axial (ax), coronal (co), and sagittal (sa) views, with four different fields-of-views (30mm: orange; 45mm: red; 85mm: green; 128mm: blue) surrounding lymph nodes.
}
\label{fig:ex_lns}
\end{figure}

\textbf{Thoracoabdominal Lymph Node Datasets.}
We use the publicly available dataset from \cite{roth2015improving,seff20142d}.
There are 388 mediastinal LNs labeled by radiologists in 90 patient CT scans, and 595 abdominal LNs in 86 patient CT scans.
To facilitate comparison, we adopt the data preparation protocol of \cite{roth2015improving}, where positive and negative LN candidates are sampled with the fields-of-view (FOVs) of 30mm to 45mm, surrounding the annotated and detected LN centers (obtained by a candidate generation process).
More precisely, \cite{roth2015improving,seff20142d,seff20152d} follow a coarse-to-fine CADe scheme, partially inspired by \cite{lu2011coarse}, which operates with $\sim100\%$ detection recalls at the cost of approximately 40 false or negative LN candidates per patient scan.
In this work, positive and negative LN candidate are first sampled up to 200 times with translations and rotations. 
Afterwards, negative LN samples are randomly re-selected at a lower rate close to the total number of positives.
LN candidates are randomly extracted from fields-of-view (FOVs) spanning 35mm to 128mm in soft-tissue window [-100, 200HU]. This allows us to capture multiple spatial scales of image context \cite{Farabet2013,mostajabi2014feedforward}).
The samples are then rescaled to a $64\times 64$ pixel resolution via B-spline interpolation. A few examples of LNs with axial, coronal, and sagittal views encoded in RGB color images \cite{roth2015improving} are shown in Figure \ref{fig:ex_lns}.

%The assumption that there are or must be pixel-wise spatial correlations among channels does not apply in the CNN model formulation. The convolutional kernels or filters in the lower-level CNN architectures can learn the optimal weights to linearly combine the evidences or observations from multiple channels as the process of dot-product. This is indeed a flexible learning representation (built-in CNN) to naturally combine multiple sources or channels of information. In recent work \cite{roth2015deeporgan}, even heterogeneous class-conditional probability maps can be combined with raw CT images to show performance benefits. This set-up is well aligned with other general computer vision work \cite{liang2015human} where heterogeneous channels of image information are fed into CNN convolutional layers for joint filtering. Last, if there are correlations among CNN input channels, it may be expected to observe the corresponding correlated patterns in the learned filters although  this is not a required condition.

Unlike the heart or the liver, lymph nodes have no pre-determined anatomic orientation. Hence, the purely random image resampling (with respect to scale, displacement and orientation) and reformatting (the axial, coronal, and sagittal views are in any system randomly resampled coordinates) is a natural choice, which also happens to yield high CNN performance.  Although we integrate three channels of information from three orthogonal views for LN detection, the pixel-wise spatial correlations between or among channels are not necessary. The convolutional kernels in the lower level CNN architectures can learn the optimal weights to linearly combine the observations from the axial, coronal, and sagittal channels by computing their dot-products.
Transforming axial, coronal, and sagittal representations to RGB also facilitates transfer learning from CNN models trained on ImageNet.

This learning representation (i.e., ``built-in CNN'') is flexible, in that it naturally combines multiple sources or channels of information. In the recent literature \cite{roth20152DeepOrgan}, even heterogeneous class-conditional probability maps can be combined with raw images to improve performance.
This set-up is similar to that of other works in computer vision, such as \cite{liang2015human}, where heterogeneous image information channels are jointly fed into the CNN convolutional layers for high-accuracy human parsing and segmentation. Finally, if there are correlations among CNN input channels, one may observe the corresponding correlated patterns in the learned filters.

In summary, the assumption that there are or must be pixel-wise spatial correlations among input channels does not apply to the CNN model representation. For other medical imaging problems, such as pulmonary embolism detection \cite{tajbakhsh2015computer}, in which orientation can be constrained along the attached vessel axis, vessel-aligned multi-planar image representation (MPR) is more effective than randomly aligned MPR.

\textbf{Interstitial Lung Disease Dataset.} We utilize the publicly available dataset of \cite{depeursinge2012building}. It contains 905 image slices from 120 patients, with six lung tissue types annotations containing at least one of the following: healthy (NM), emphysema (EM), ground glass (GG), fibrosis (FB), micronodules (MN) and consolidation (CD) (Figure \ref{fig:ild_patch_examples}).
At the slice level, the objective is to classify the status of ``presence/absence'' of any of the six ILD classes for an input axial CT slice \cite{gao2014holistic}.
Characterizing an arbitrary CT slice against any possible ILD type, without any manual ROI (in contrast to \cite{song2013feature,song2015large}), can be useful for large-scale patient screening.
For slice-level ILD classification, we sampled the slices 12 times with random translations and rotations.
After this, we balanced the numbers of CT slice samples for the six classes by randomly sampling several instances at various rates.
For patch-based classification, we sampled up to 100 patches of size $64\times 64$ from each ROI.
This dataset is divided into five folds with disjoint patient subsets.
The average number of CT slices (training instances) per fold is small, as shown in Table \ref{tab:ild_num_classes_5fold}.
Slice-level ILD classification is a very challenging task where CNN models need to learn from very small numbers of training examples and predict ILD labels on unseen patients.

In the publicly available ILD dataset, very few CT slices are labeled as normal or healthy. The remaining CT slices cannot be simply classified as normal, because many ILD disease regions or slices have not yet been labeled. ILD \cite{depeursinge2012building} is a partially labeled database; this is one of its main limitations. Research is being conducted to address this issue. In particular,\cite{gao2016isbi} has proposed to fully label the ILD dataset pixel-wise via proposed segmentation label propagation.

\begin{figure}[t]
\begin{center}
   \includegraphics[width=1\linewidth]{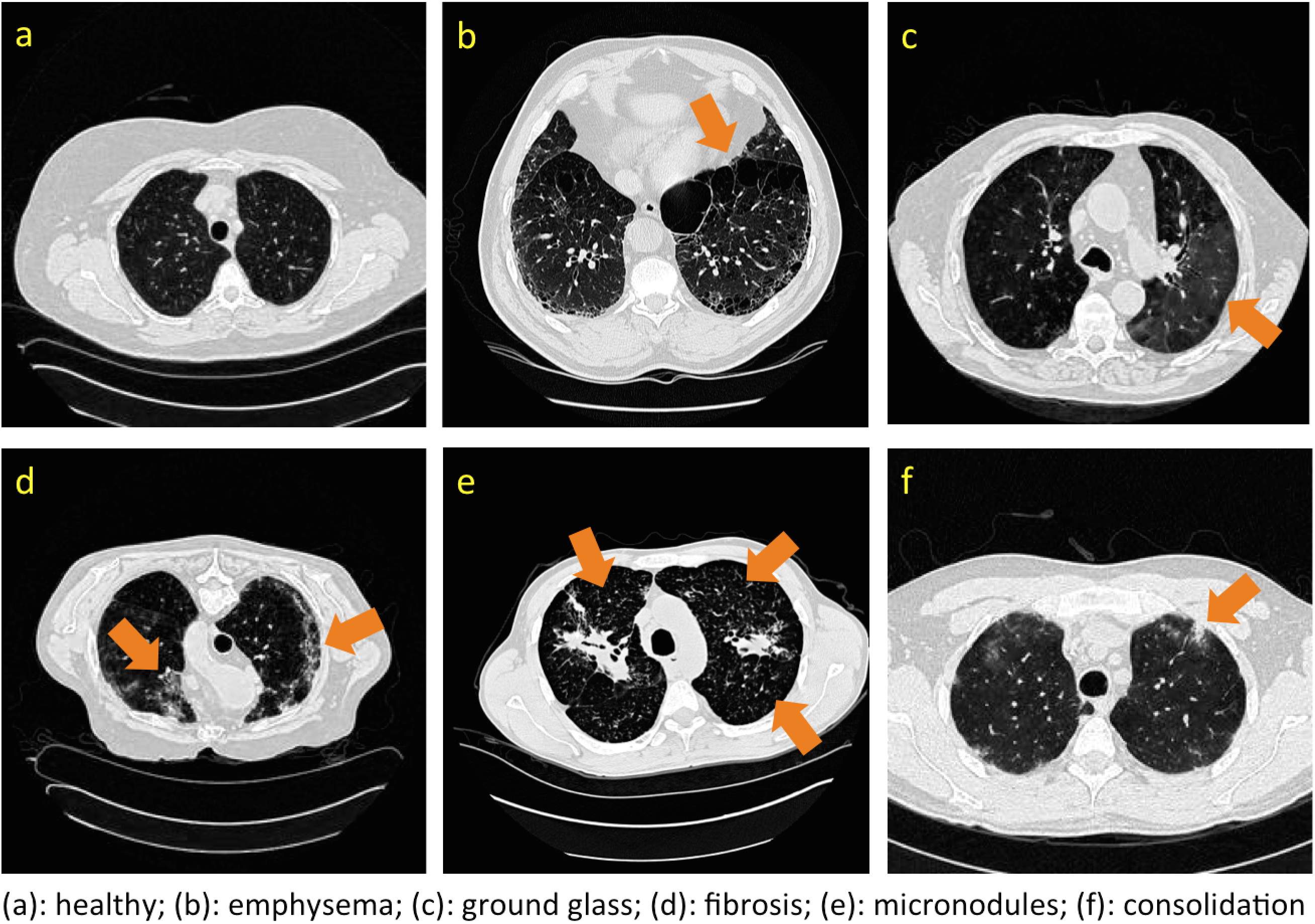}
\end{center}
   \caption{Some examples of CT image slices with six lung tissue types in the ILD dataset \cite{depeursinge2012building}. Disease tissue types are located with dark orange arrows.}
\label{fig:ex_ilds}
\end{figure}

\begin{figure}[t]
\begin{center}
   \includegraphics[width=1\linewidth]{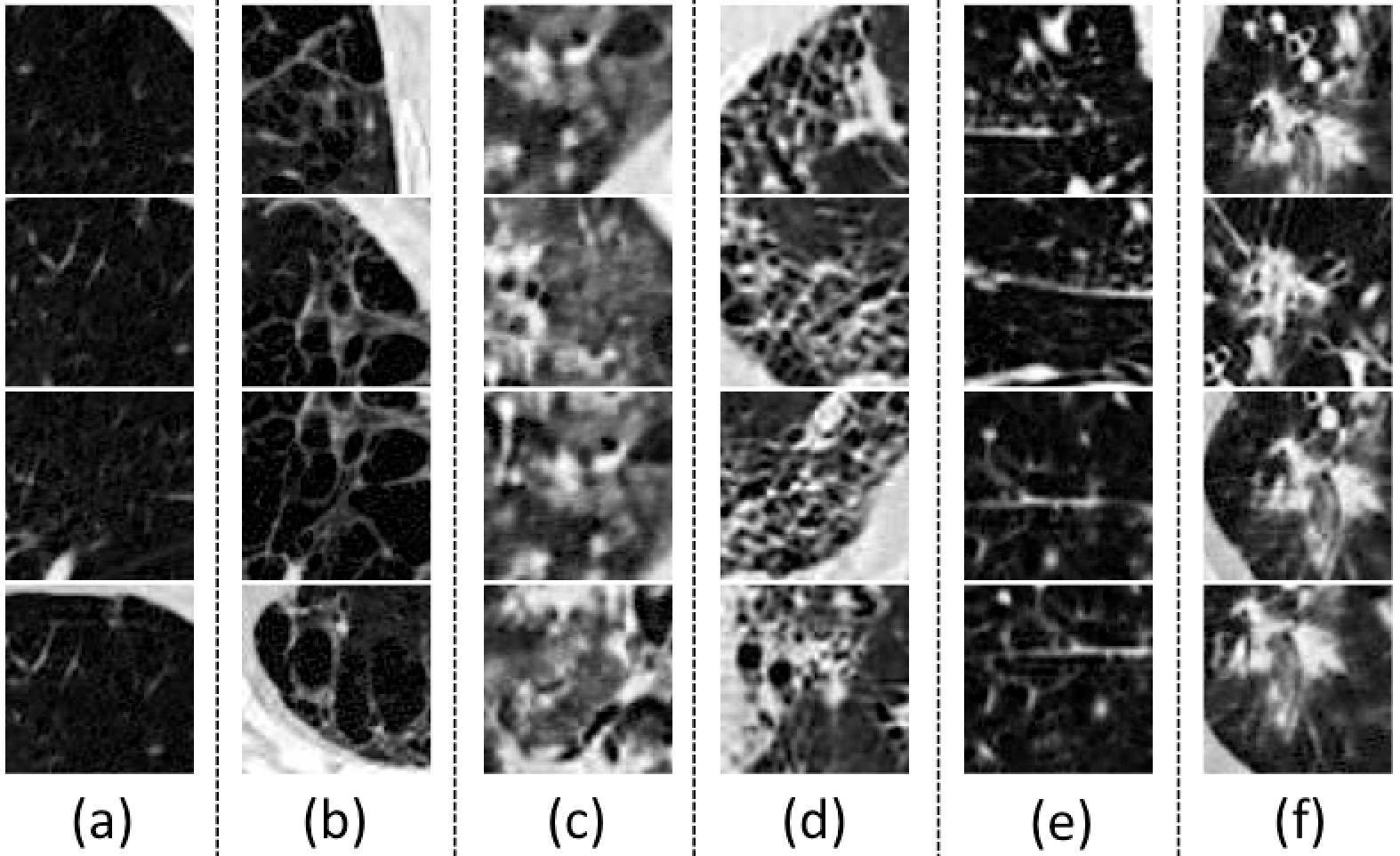}
\end{center}
   \caption{Some examples of $64 \times 64$ pixel CT image patches for (a) NM, (b) EM, (c) GG, (d) FB, (e) MN (f) CD.}
\label{fig:ild_patch_examples}
\end{figure}

\begin{table}[t]
\begin{center}
\resizebox{1\linewidth}{!}{
\begin{tabular}{|c c c c c c|}
\hline
                           normal  &    emphysema   &   ground glass   &    fibrosis  &    micronodules    &    consolidation   \\
\hline                                                                              
                          30.2 &  20.2   &  85.4  &  96.8  &  63.2    &   39.2  \\
\hline
\end{tabular}
}
\end{center}
\caption{Average number of images in each fold for disease classes, when dividing the dataset in 5-fold patient sets.}
\label{tab:ild_num_classes_5fold}
\end{table}

\begin{figure}[t]
\begin{center}
   \includegraphics[width=1\linewidth]{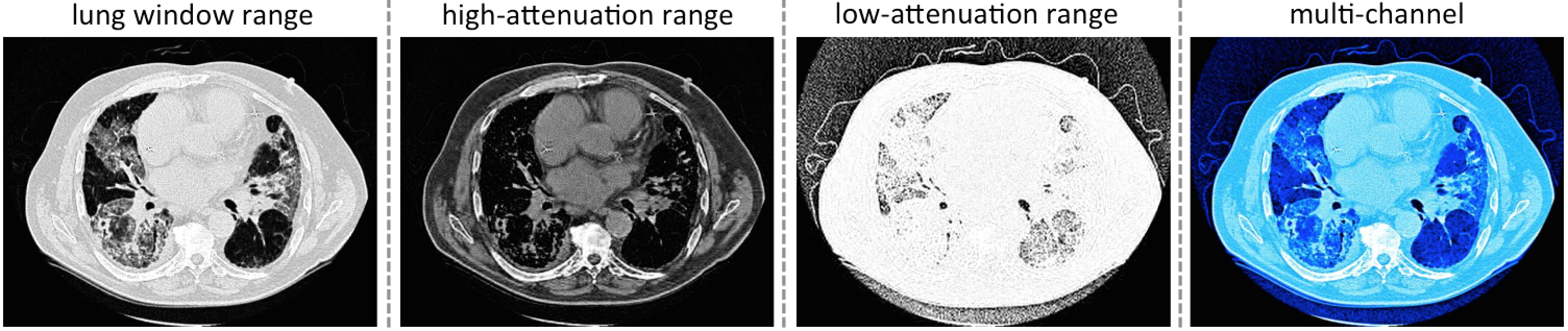}
\end{center}
   \caption{An example of lung/high-attenuation/low-attenuation CT windowing for an axis lung CT slice. We encode the lung/high-attenuation/low-attenuation CT windowing into red/green/blue channels.}
\label{fig:ex_lung_seg_rgb}
\end{figure}

To leverage the CNN architectures designed for color images and to transfer CNN parameters pre-trained on ImageNet, we transform all gray-scale axial CT slice images via three CT window ranges: lung window range [-1400, -200HU], high-attenuation range [-160, 240HU], and low-attenuation range [-1400; -950HU]. We then encode the transformed images into RGB channels (to be aligned with the input channels of CNN models \cite{krizhevsky2012imagenet,szegedy2014going} pre-trained from natural image datasets \cite{deng2009imagenet}). The low-attenuation CT window is useful for visualizing certain texture patterns of lung diseases (especially emphysema). %Since the high-attenuation range does not effectively highlight lung disease patterns, we replace it with a lung windowing performed inside the lung region. 
The usage of different CT attenuation channels improves classification results over the usage of a single CT windowing channel, as demonstrated in \cite{gao2014holistic}. More importantly, these CT windowing processes do not depend on the lung segmentation, which instead is directly defined in the CT HU space. Figure \ref{fig:ex_lung_seg_rgb} shows a representative example of lung, high-attenuation, and low-attenuation CT windowing for an axis lung CT slice.

As observed in \cite{gao2014holistic}, lung segmentation is crucial to holistic slice-level ILD classification. We empirically compare performance in two scenarios with a rough lung segmentation\footnote{This can be achieved by segmenting the lung using simple label-fusion methods \cite{wang2013multi}. In the first case, we overlay the target image slice with the average lung mask among the training folds. In the second, we perform simple morphology operations to obtain the lung boundary. In order to retain information from the inside of the lung, we apply Gaussian smoothing to the regions outside of the lung boundary.} There is no significant difference between two setups. Due to the high precision of CNN based image processing, highly accurate lung segmentation is not necessary . The localization of ILD regions within the lung is simultaneously learned through selectively weighted CNN reception fields in the deepest convolutional layers during the classification based CNN training \cite{oquab2015object,oquab2014learning}.
Some areas outside of the lung appear in both healthy or diseased images. CNN training learns to ignore them by setting very small filter weights around the corresponding regions (Figure \ref{fig:viz_ild_ex}). This observation is validated by \cite{gao2014holistic}.

\section{Methods}

In this study, we explore, evaluate and analyze the influence of various CNN Architectures, dataset characteristics (when we need more training data or better models for object detection \cite{Zhu2012do}) and CNN transfer learning from non-medical to medical image domains.
These three key elements of building effective deep CNN models for CADe problems are described below.

\subsection{Convolutional Neural Network Architectures}

We mainly explore three convolutional neural network architectures (CifarNet \cite{krizhevsky2009learning,roth2015improving}, AlexNet \cite{krizhevsky2012imagenet} and GoogLeNet \cite{szegedy2014going}) with different model training parameter values.
The current deep learning models \cite{roth2015improving,Ciresan2013,zhang2015deep} in medical image tasks are at least $2\sim5$ orders of magnitude smaller than even AlexNet \cite{krizhevsky2012imagenet}.
More complex CNN models \cite{roth2015improving,Ciresan2013} have only about 150K or 15K parameters.
Roth et al. \cite{roth2015improving} adopt the CNN architecture tailored to the Cifar-10 dataset \cite{krizhevsky2009learning} and operate on image windows of $32\times32\times3$ pixels for lymph node detection, while the simplest CNN in \cite{li2014medical} has only one convolutional, pooling, and FC layer, respectively. 

\begin{figure*}[t]
\begin{center}
   \includegraphics[width=1\linewidth]{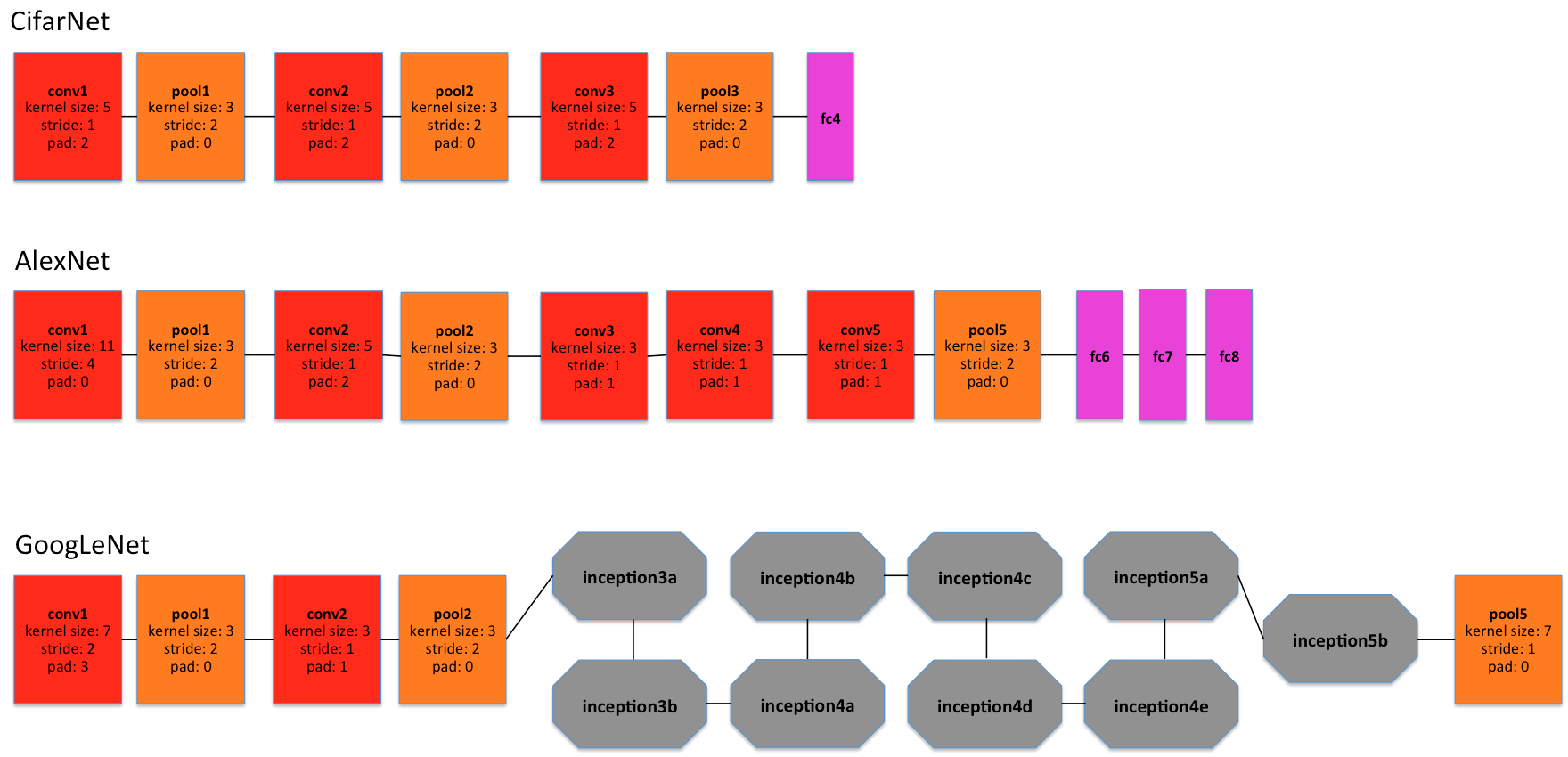}
\end{center}
   \caption{A simplified illustration of the CNN architectures used. GoogLeNet \cite{szegedy2014going} contains two convolution layers, three pooling layers, and nine inception layers. Each of the inception layer of GoogLeNet consists of six convolution layers and one pooling layer.}
\label{fig:cnn_architectures}
\end{figure*}

We use CifarNet \cite{krizhevsky2009learning} as used in \cite{roth2015improving} as a baseline for the LN detection.
AlexNet \cite{krizhevsky2012imagenet} and GoogLeNet \cite{szegedy2014going} are also modified to evaluate these state-of-the-art CNN architecture from ImageNet classification task \cite{Russakovsky2014ILSVRC} to our CADe problems and datasets.
A simplified illustration of three CNN architectures exploited is shown in Figure \ref{fig:cnn_architectures}.
CifarNet always takes $32\times 32\times 3$ image patches as input while AlexNet and GoogLeNet are originally designed for the fixed image dimension of $256\times 256\times 3$ pixels.
We also reduced the filter size, stride and pooling parameters of AlexNet and GoogLeNet to accommodate a smaller input size of $64\times 64\times 3$ pixels.
We do so to produce and evaluate ``simplified'' AlexNet and GoogLeNet versions that are better suited to the smaller scale training datasets common in CADe problems.
Throughout the paper, we refer to the models as CifarNet (32x32) or CifarNet (dropping 32x32); AlexNet (256x256) or AlexNet-H (high resolution); AlexNet (64x64) or AlexNet-L (low resolution); GoogLeNet (256x256) or GoogLeNet-H and GoogLeNet (64x64) or GoogLeNet-L (dropping 3 since all image inputs are three channels).

\begin{figure}[t]
\begin{center}
   \includegraphics[width=.8\linewidth]{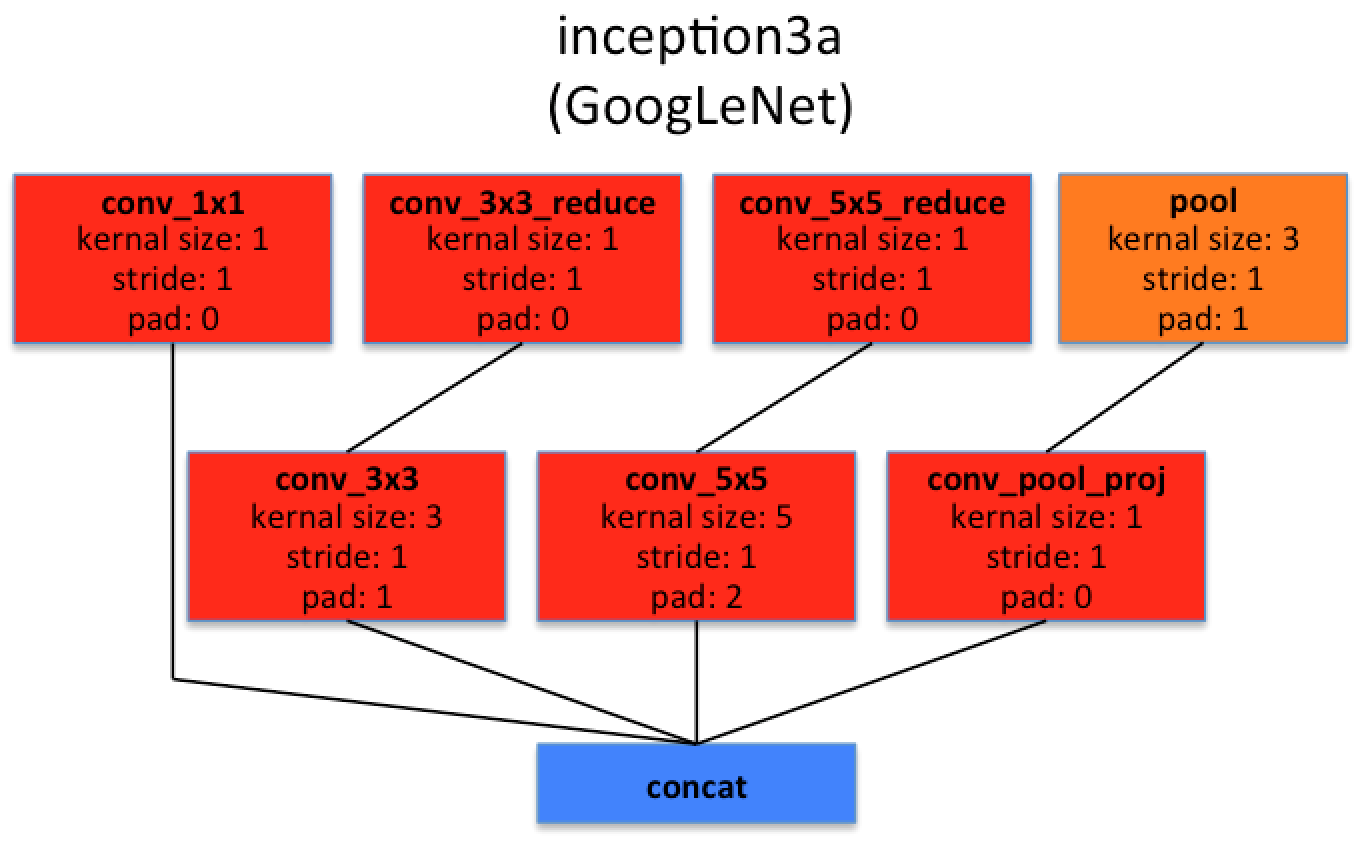}
\end{center}
   \caption{Illustration of \texttt{inception3a} layer of GoogLeNet. Inception layers of GoogLeNet consist of six convolution layers with different kernel sizes and one pooling layer.}
\label{fig:cnn_architectures_inception}
\end{figure}

\begin{figure}[t]
\begin{center}
   \includegraphics[width=1\linewidth]{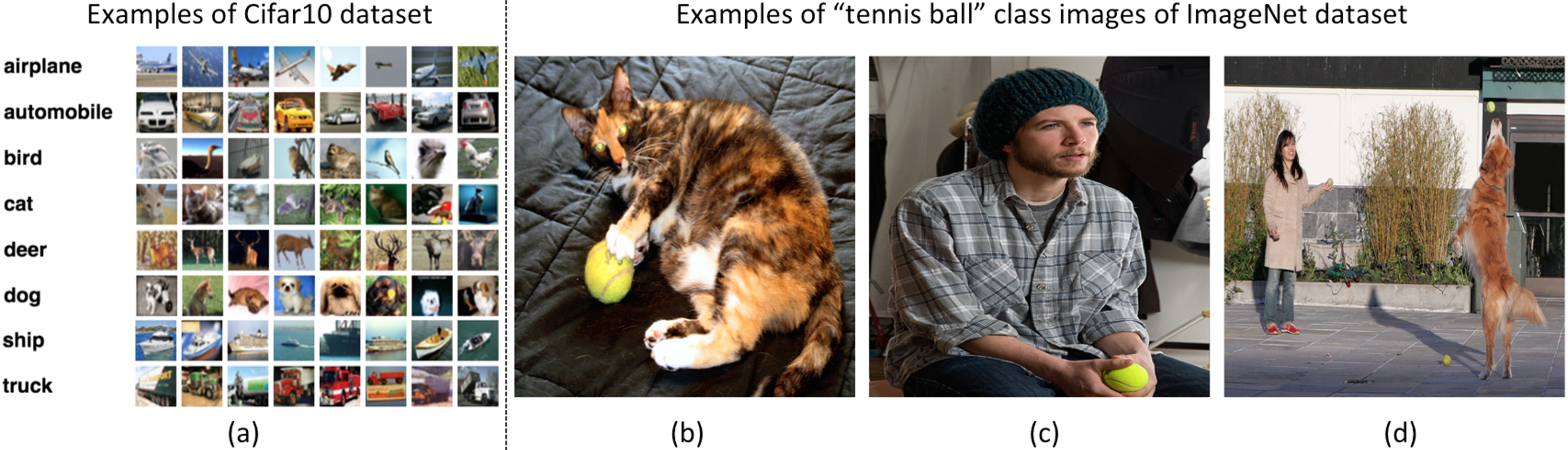}
\end{center}
   \caption{Some examples of Cifar10 dataset and some images of ``tennis ball'' class from ImageNet dataset. Images of Cifar10 dataset are small ($32\times 32$) images with object of the image class category in the center. Images of ImageNet dataset are larger ($256\times 256$), where object of the image class category can be small, obscure, partial, and sometimes in a cluttered environment.}
\label{fig:cifar10_imagenet_examples}
\end{figure}

\paragraph{CifarNet}
CifarNet, introduced in \cite{krizhevsky2009learning}, was the state-of-the-art model for object recognition on the Cifar10 dataset, which consists of $32\times 32$ images of 10 object classes.
The objects are normally centered in the images.
Some example images and class categories from the Cifar10 dataset are shown in Figure \ref{fig:cifar10_imagenet_examples}.
CifarNet has three convolution layers, three pooling layers, and one fully-connected layer.
This CNN architecture, also used in \cite{roth2015improving} has about 0.15 million free parameters.
We adopt it as a baseline model for the LN detection.

\paragraph{AlexNet}
The AlexNet architecture was published in \cite{krizhevsky2012imagenet}, achieved significantly improved performance over the other non-deep learning methods for ImageNet Large Scale Visual Recognition Challenge (ILSVRC) 2012.
This success has revived the interest in CNNs \cite{LeCun1998gradient} in computer vision.
ImageNet consists of 1.2 million $256\times 256$ images belonging to 1000 categories.
At times, the objects in the image are small and obscure, and thus pose more challenges for learning a successful classification model.
More details about the ImageNet dataset will be discussed in Sec. \ref{sec:imagenet}.
AlexNet has five convolution layers, three pooling layers, and two fully-connected layers with approximately 60 million free parameters.
AlexNet is our default CNN architecture for evaluation and analysis in the remainder of the paper.

\paragraph{GoogLeNet}
The GoogLeNet model proposed in \cite{szegedy2014going}, is significantly more complex and deep than all previous CNN architectures.
More importantly, it also introduces a new module called ``Inception'', which concatenates filters of different sizes and dimensions into a single new filter (refer to Figure \ref{fig:cnn_architectures_inception}).
Overall, GoogLeNet has two convolution layers, two pooling layers, and nine ``Inception'' layers.
Each ``Inception'' layer consists of six convolution layers and one pooling layer.
An illustration of an ``Inception'' layer (\texttt{inception3a}) from GoogLeNet is shown in Figure \ref{fig:cnn_architectures_inception}.
GoogLeNet is the current state-of-the-art CNN architecture for the ILSVRC challenge, where it achieved 5.5\% top-5 classification error on the ImageNet challenge, compared to AlexNet's 15.3\% top-5 classification error.

\subsection{ImageNet: Large Scale Annotated Natural Image Dataset}
\label{sec:imagenet}

ImageNet \cite{deng2009imagenet} has more than 1.2 million $256\times 256$ images categorized under 1000 object class categories.
There are more than 1000 training images per class.
The database is organized according to the WordNet \cite{miller1995wordnet} hierarchy, which currently contains only nouns in 1000 object categories.
The image-object labels are obtained largely through crowd-sourcing, e.g., Amazon Mechanical Turk, and human inspection.
Some examples of object categories in ImageNet are ``sea snake'', ``sandwich'', ``vase'', ``leopard'', etc.
ImageNet is currently the largest image dataset among other standard datasets for visual recognition.
Indeed, the Caltech101, Caltech256 and Cifar10 dataset merely contain 60000 $32\times 32$ images and 10 object classes.
Furthermore, due to the large number (1000+) of object classes, the objects belonging to each ImageNet class category can be occluded, partial and small, relative to those in the previous public image datasets.
This significant intra-class variation poses greater challenges to any data-driven learning system that builds a classifier to fit given data and generalize to unseen data.
For comparison, some example images of Cifar10 dataset and ImageNet images in the ``tennis ball'' class category are shown in Figure \ref{fig:cifar10_imagenet_examples}.
The ImageNet dataset is publicly available, and the ImageNet Large Scale Visual Recognition Challenge (ILSVRC) has become the standard benchmark for large-scale object recognition.

\subsection{Training Protocols and Transfer Learning}

When {\bf learned from scratch}, all the parameters of CNN models are initialized with random Gaussian distributions and trained for 30 epochs with the mini-batch size of 50 image instances.
Training convergence can be observed within 30 epochs. The other hyperparameters are momentum: 0.9; weight decay: 0.0005; (base) learning rate: 0.01, decreased by a factor of 10 at every 10 epochs. We use the Caffe framework \cite{jia2013caffe} and NVidia K40 GPUs to train the CNNs. 

AlexNet and GoogLeNet CNN models can be either learned from scratch or {\bf fine-tuned from pre-trained models}. Girshick et al. \cite{Girshick15} find that, by applying ImageNet pre-trained ALexNet to PASCAL dataset \cite{Everingham2015Pascal}, performances of semantic 20-class object detection and segmentation tasks significantly improve over previous methods that use no deep CNNs. AlexNet can be fine-tuned on the PASCAL dataset to surpass the performance of the ImageNet pre-trained AlexNet, although the difference is not as significant as that between the CNN and non-CNN methods. Similarly, \cite{razavian2014cnn,zhou2014learning} also demonstrate that better performing deep models are learned via CNN transfer learning from ImageNet to other datasets of limited scales.

Our hypothesis on CNN parameter transfer learning is the following: despite the disparity between natural images and natural images, CNNs comprehensively trained on the large scale well-annotated ImageNet may still be transferred to make medical image recognition tasks more effective. Collecting and annotating large numbers of medical images still poses significant challenges. On the other hand, the mainstream deep CNN architectures (e.g., AlexNet and GoogLeNet) contain tens of millions of free parameters to train, and thus require sufficiently large numbers of labeled medical images. 

For transfer learning, we follow the approach of \cite{razavian2014cnn,Girshick15} where all CNN layers except the last are fine-tuned at a learning rate 10 times smaller than the default learning rate. The last fully-connected layer is random initialized and freshly trained, in order to accommodate the new object categories in our CADe applications. Its learning rate is kept at the original 0.01. We denote the models with random initialization or transfer learning as AlexNet-RI and AlexNet-TL, and GoogLeNet-RI and GoogLeNet-TL. We found that the transfer learning strategy yields the best performance results. Determining the optimal learning rate for different layers is challenging, especially for very deep networks such as GoogLeNet.

We also perform experiments using {\bf ``off-the-shelf''} CNN features of AlexNet pre-trained on ImageNet and training only the final classifier layer to complete the new CADe classification tasks. Parameters in the convolutional and fully connected layers are fixed and are used as deep image extractors, as in \cite{Bar2015,Ginneken15,Ciompi2015automatic}. We refer to this model as AlexNet-ImNet in the remainder of the paper. Note that \cite{Bar2015,Ginneken15,Ciompi2015automatic} train support vector machines and random forest classifiers using ImageNet pre-trained CNN features.  
Our simplified implementation is intended to determine whether fine-tuning the ``end-to-end'' CNN network is necessary to improve performance, as opposed to merely training the final classification layer. This is a slight modification from the method described in \cite{Bar2015,Ginneken15,Ciompi2015automatic}.

Finally, transfer learning in CNN representation, as empirically verified in previous literature \cite{Gupta2014Learning,Gupta2015Indoor,Gupta2013Natural,Shin2015Interleaved,Chen2015Automatic}, can be effective in various cross-modality imaging settings (RGB images to depth images \cite{Gupta2014Learning,Gupta2015Indoor}, natural images to general CT and MRI images \cite{Shin2015Interleaved}, and natural images to neuroimaging \cite{Gupta2013Natural} or ultrasound \cite{Chen2015Automatic} data). More thorough theoretical studies on cross-modality imaging statistics and transferability will be needed for future studies.

\section{Evaluations and Discussions}

In this section, we evaluate and compare the performances of nine CNN model configurations (CifarNet, AlexNet-ImNet,
AlexNet-RI-H, AlexNet-TL-H, AlexNet-RI-L, GoogLeNet-RI-H, GoogLeNet-TL-H, GoogLeNet-RI-L and combined) on two important CADe problems using publicly available datasets \cite{roth2015improving,seff20142d,depeursinge2012building}.

\subsection{Thoracoabdominal Lymph Node Detection}
\label{sec:LN_detection_results}

\begin{figure*}[t]
\begin{center}
   \includegraphics[width=.96\linewidth]{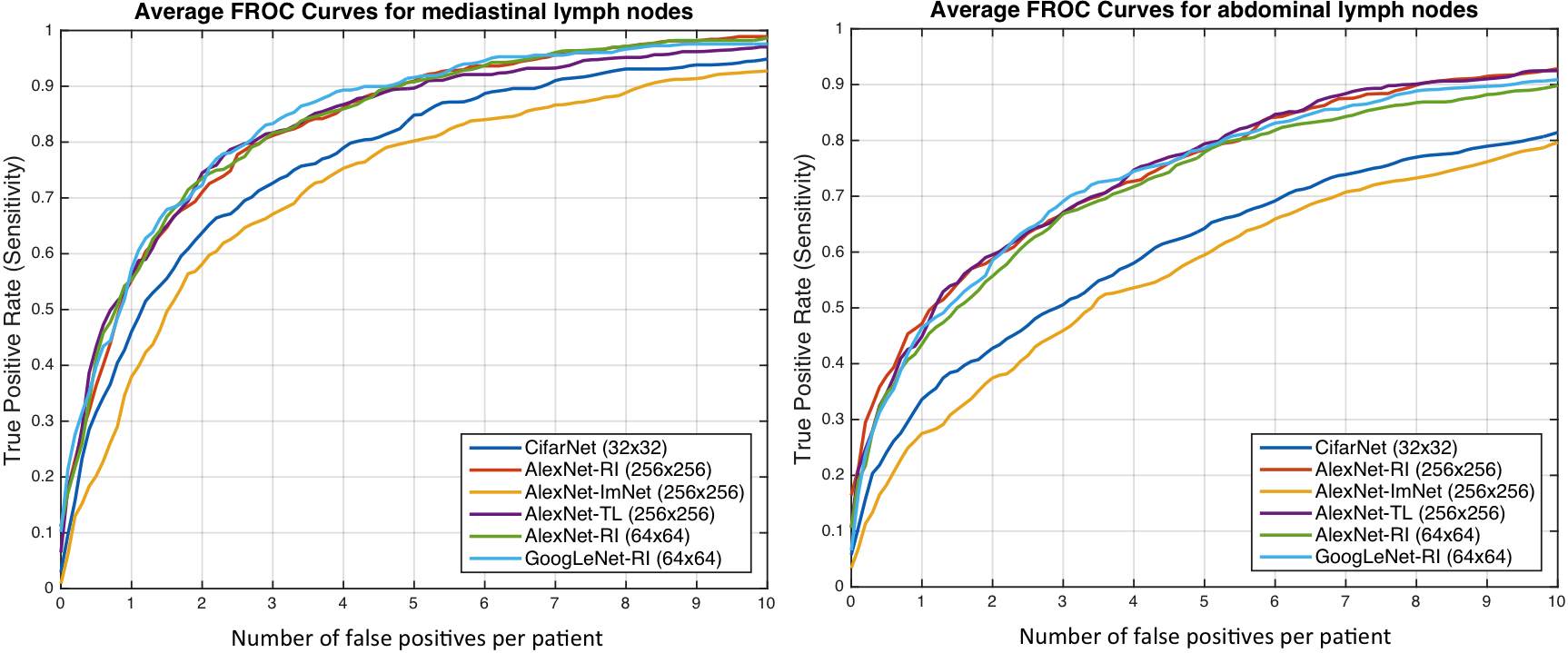}
\end{center}
   \caption{FROC curves averaged on three-fold CV for the abdominal (left) and mediastinal (right) lymph nodes using different CNN models.}
\label{fig:ild_conv1s}
\end{figure*}

We train and evaluate CNNs using three-fold cross-validation (folds are split into disjoint sets of patients), with the different CNN architectures described above.
In testing, each LN candidate has multiple random 2.5D views tested by CNN classifiers to generate LN class probability scores.
We follow the random view aggregation by averaging probabilities, as in \cite{roth2015improving}. 

We first sample the LN image patches at a $64\times 64$ pixel resolution. We then up-sample the $64\times 64$ pixel LN images via bi-linear interpolation to $256\times 256$ pixels, in order to accommodate AlexNet-RI-L, AlexNet-TL-H, GoogLeNet-RI-H and GoogLeNet-TL-H. For the modified AlexNet-RI-L at ($64\times 64$) pixel resolution, we reduce the number of first layer convolution filters from 96 to 64 and reduce the stride from 4 to 2. For the modified GoogLeNet-RI ($64\times 64$), we decrease the number of first layer convolution filters from 64 to 32, the pad size from 3 to 2, the kernel size from 7 to 5, stride from 2 to 1 and the stride of the subsequent pooling layer from 2 to 1. We slightly reduce the number of convolutional filters in order to accommodate the smaller input image sizes of target medical image datasets \cite{roth2015improving,depeursinge2012building}, while preventing over-fitting. This eventually improves performance on patch-based classification. CifarNet is used in \cite{roth2015improving} to detect LN samples of $32\times 32\times 3$ images. For consistency purposes, we down-sample $64\times 64\times 3$ resolution LN sample images to the dimension of $32\times 32\times 3$.

\begin{table}
\begin{center}
\resizebox{1.\linewidth}{!}{
\begin{tabular}{|c|| c c | c c |}
\hline
Region              &     \multicolumn{2}{|c|}{Mediastinum}   &    \multicolumn{2}{|c|}{Abdomen}  \\
\hline
Method              &     AUC     &    TPR/3FP            &     AUC     &    TPR/3FP     \\
\hline\hline
\cite{seff20142d}   &      -    &      0.63             &      -    &      0.70      \\
\cite{roth2015improving}  &     \textbf{0.92}    &      0.70             &     \textbf{0.94}    &      \textbf{0.83}      \\
\cite{seff20152d}   &      -    &      \textbf{0.78}             &      -    &      0.78      \\
\hline
CifarNet    &     0.91    &      0.70             &     0.81    &      0.44      \\
AlexNet-ImNet &  0.89    &      0.63             &     0.80    &      0.41      \\
AlexNet-RI-H &     0.94    &      0.79             &     0.92    &      0.67      \\
AlexNet-TL-H &     0.94    &      0.81             &     0.92    &      0.69      \\
\hline
GoogLeNet-RI-H &     0.85 & 0.61 & 0.80 & 0.48      \\
GoogLeNet-TL-H &     0.94    &      0.81             &     \textbf{0.92}    &      \textbf{0.70}      \\
\hline
AlexNet-RI-L &     0.94    &      0.77               &     0.88    &      0.61      \\
GoogLeNet-RI-L &     \textbf{0.95}  &     \textbf{0.85}    &     0.91    &      0.69      \\
\hline
Combined &  0.95    &      0.85             &     0.93    &      0.70      \\
\hline
\end{tabular}
}
\end{center}
\caption{Comparison of mediastinal and abdominal LN detection results using various CNN models. Numbers in bold indicate the best performance values on classification accuracy.
}
\label{tab:ln_eval_comparison}
\end{table}

Results for lymph node detection in the mediastinum and abdomen are reported in Table \ref{tab:ln_eval_comparison}. 
FROC curves are illustrated in Figure \ref{fig:ild_conv1s}.
The area-under-the-FROC-curve (AUC) and true positive rate (TPR, recall or sensitivity) at three false positives per patient (TPR/3FP) are used as performance metrics.
Of the nine investigated CNN models, CifarNet, AlexNet-ImNet and GoogLeNet-RI-H generally yielded the least competitive detection accuracy results.
Our LN datasets are significantly more complex (i.e., display much larger within-class appearance variations), especially due to the extracted fields-of-view (FOVs) of (35mm-128mm) compared to (30mm-45mm) in \cite{roth2015improving}, where CifarNet is also employed.
In this experiment, CifarNet is under-trained with respect to our enhanced LN datasets, due to its limited input resolution and parameter complexity.
The inferior performance of AlexNet-ImNet implies that using the pre-trained ImageNet CNNs alone as ``off-the-shelf'' deep image feature extractors may not be optimal or adequate for mediastinal and abdominal LN detection tasks. 
To complement ``off-the-shelf'' CNN features, \cite{Bar2015,Ginneken15,Ciompi2015automatic} all add and integrate various other hand-crafted image features as hybrid inputs for the final CADe classification.

GoogLeNet-RI-H performs poorly, as it is susceptible to over-fitting. No sufficient data samples are available to train GoogLeNet-RI-H with random initialization.
Indeed, due to GoogLeNet-RI-H's complexity and 22-layer depth, million-image datasets may be required to properly train this model.
However, GoogLeNet-TL-H significantly improves upon GoogLeNet-RI-H (0.81 versus 0.61 TPR/3FP in mediastinum; 0.70 versus 0.48 TPR/3FP in abdomen). This indicates that transfer learning offers a much better initialization of CNN parameters than random initialization. Likewise, AlexNet-TL-H consistently outperforms AlexNet-RI-H, though by smaller margins (0.81 versus 0.79 TPR/3FP in mediastinum; 0.69 versus 0.67 TPR/3FP in abdomen). This is also consistent with the findings reported for ILD detection in Table \ref{tab:ild_accuracies} and Figure \ref{fig:ild_alexnet_training_traces}.

GoogLeNet-TL-H yields results similar to AlexNet-TL-H's for the mediastinal LN detection, and slightly outperforms Alex-Net-H for abdominal LN detection.
AlexNet-RI-H exhibits less severe over-fitting than GoogLeNet-RI-H.
We also evaluate a simple ensemble by averaging the probability scores from five CNNs: AlexNet-RI-H, AlexNet-TL-H, AlexNet-RI-H, GoogLeNet-TL-H and GoogLeNet-RI-L.
This combined ensemble outputs the classification accuracies matching or slightly exceeding the best performing individual CNN models on the mediastinal or abdominal LN detection tasks, respectively.

Many of our CNN models achieve notably better (FROC-AUC and TPR/3FP) results than the previous state-of-the-art models \cite{seff20152d} for {\bf mediastinal} LN detection: GoogLeNet-RI-L obtains an AUC=0.95 and 0.85 TPR/3FP, versus AUC=0.92 and 0.70 TPR/3FP \cite{roth2015improving} and 0.78 TPR/3FP \cite{seff20152d} which uses stacked shallow learning.
This difference lies in the fact that annotated lymph node segmentation masks are required  to learn a mid-level semantic boundary detector \cite{seff20152d}, whereas CNN approaches only need LN locations for training \cite{roth2015improving}. In {\bf abdominal} LN detection, \cite{roth2015improving} obtains the best trade-off between its CNN model complexity and sampled data configuration. Our best performing CNN model is GoogLeNet-TL (256x256) which obtains an AUC=0.92 and 0.70 TPR/3FP.
% The architecture of GoogLeNet-RI-L is far less complex than GoogLeNet–TL-H's.
% Its input image dimensions are reduced from $256\times 256$ to $64\times 64$, and it contains less free parameters. However, it can no longer leverage on transfer learning from ImageNet \cite{szegedy2014going}.

The main difference between our dataset preparation protocol and that from \cite{roth2015improving} is a more aggressive extraction of random views within a much larger range of FOVs.
The usage of larger FOVs to capture more image spatial context is inspired by deep zoom-out features \cite{mostajabi2014feedforward} that improve semantic segmentation. This image sampling scheme contributes to our best reported performance results in both mediastinal LN detection (in this paper) and automated pancreas segmentation \cite{roth20152DeepOrgan}. As shown in Figure \ref{fig:ex_lns}, abdominal LNs are surrounded by many other similar looking objects.
Meanwhile, mediastinal LNs are more easily distinguishable, due to the images' larger spatial contexts.
Finally, from the perspective of the data-model trade-off: ``{\em Do We Need More Training Data or Better Models?}'' \cite{Zhu2012do}, more abdomen CT scans from distinct patient populations need to be acquired and annotated, in order to take full advantage of deep CNN models of high capacity.
Nevertheless, deeper and wider CNN models (e.g., GoogLeNet-RI-L and GoogLeNet-TL-H versus Cifar-10 \cite{roth2015improving}) have shown improved results in the mediastinal LN detection.

\begin{figure}[t]
\begin{center}
   \includegraphics[width=1\linewidth]{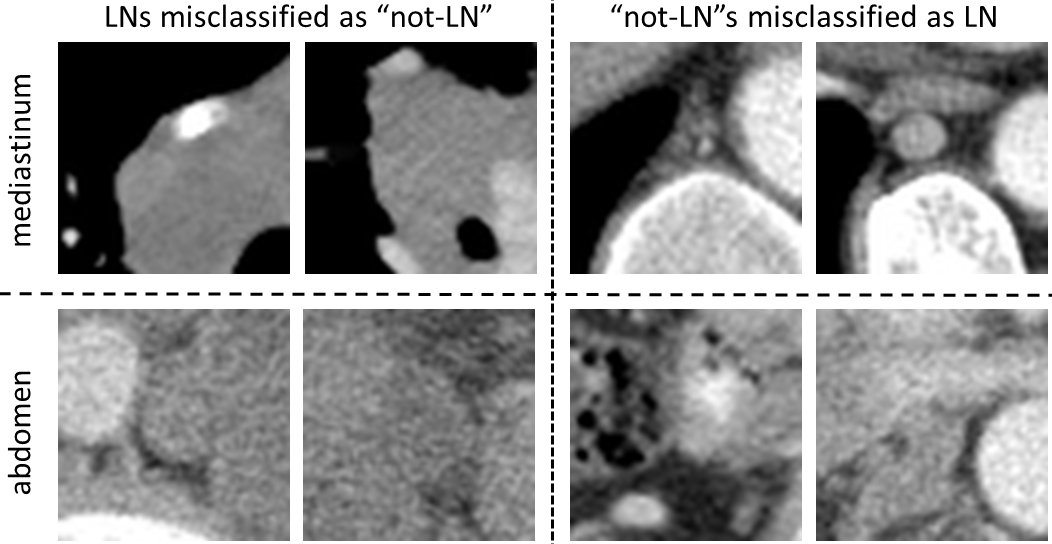}
\end{center}
   \caption{Examples of misclassified lymph nodes (in axial view) of both false negatives ({\bf Left}) and false positives  ({\bf Right}). Mediastinal LN examples are shown in the {\bf upper} row, and abdominal LN examples in the {\bf bottom} row.}
\label{fig:miscls_exs_lns}
\end{figure}

Figure \ref{fig:miscls_exs_lns} provides examples of misclassified lymph nodes (in axial view) (both false negatives ({\bf Left}) and false positives({\bf Right})), from the Abdomen and Mediastinum datasets. The overall reported LN detection results are clinically significant, as indicated in \cite{lkim2014rsna}.

\begin{table*}
\begin{center}
\resizebox{0.8\linewidth}{!}{
\begin{tabular}{|c| c c c| c c | c |}
\hline
Method        &  AlexNet-ImNet & AlexNet-RI & AlexNet-TL     & GoogLeNet-RI &   GoogLeNet-TL & Avg-All \\
\hline
% Slice-level   &      0.89    &     0.75       & \textbf{0.91}  &     0.79     &        0.90      \\
Slice-CV5 &      0.45    &     0.44       &    0.46        &     0.41     &  \textbf{0.57}  & 0.53    \\
Patch-CV5 &      0.76    &     0.74       &    0.76        &     0.75     &  0.76    & \textbf{0.79}   \\
\hline
\end{tabular}
}
\end{center}
\caption{Comparison of interstitial lung disease classification accuracies on both slice-level (Slice-CV5) and patch-based (Patch-CV5) classification using five-fold CV. Bold numbers indicate the best performance values on classification accuracy.}
\label{tab:ild_accuracies}
\end{table*}

\subsection{Interstitial Lung Disease Classification} \label{sec:ILD_detection_results}

The CNN models evaluated in this experiment are \textit{1)} AlexNet-RI (training from scratch on the ILD dataset with random initialization); \textit{2)} AlexNet-TL (with transfer learning from \cite{krizhevsky2012imagenet}); \textit{3)} AlexNet-ImNet: pre-trained ImageNet-CNN model \cite{krizhevsky2012imagenet} with only the last cost function layer retrained from random initialization, according to the six ILD classes (similar to \cite{Ginneken15} but without using additional hand-crafted non-deep feature descriptors, such as GIST and BoVW); \textit{4)} GoogLeNet-RI (random initialization); \textit{5)} GoogLeNet-TL (GoogLeNet with transfer learning from \cite{szegedy2014going}). All ILD images (patches of $64\times 64$ and CT axial slices of $512\times 512$) are re-sampled to a fixed dimension of $256\times 256$ pixels.

\begin{table}
\begin{center}
\resizebox{1\linewidth}{!}{
\begin{tabular}{|c|c c c c c c|}
\hline
                        &   NM  &    EM   &   GG   &    FB  &    MN    &    CD   \\
\hline                                                                              
Patch-LOO \cite{song2013feature}  & 0.84  &  0.75   &  0.78  &  0.84  &  0.86    &    -    \\
Patch-LOO \cite{song2015large}    & 0.88  &  0.77   &  0.80  &  0.87  &  0.89    &    -    \\
\hline
Patch-CV10 \cite{li2014medical}    & 0.84  &  0.55   &  0.72  &  0.76  &  0.91    &    -    \\
Patch-CV5           & 0.64  &  0.81   &  0.74  &  0.78  &  0.82    &  0.64   \\
\hline \hline
Slice-Test \cite{gao2014holistic}  & 0.40  &  1.00   &  0.75  &  0.80  &  0.56    &  0.50   \\
\hline                                                                              
Slice-CV5           & 0.22  &  0.35   &  0.56  &  0.75  &  0.71    &  0.16   \\
Slice-Random          & 0.90  &  0.86   &  0.85  &  0.94  &  0.98    &  0.83   \\
\hline
\end{tabular} 
}
\end{center}
\caption{Comparison of interstitial lung disease classification results using F-scores: NM, EM, GG, FB, MN and CD. }
\label{tab:ild_f1_scores}
\end{table}

\begin{table}
\begin{center}
\resizebox{.95\linewidth}{!}{
\begin{tabular}{|c|c c c c c c|}
\hline
Ground       &  \multicolumn{6}{|c|}{Prediction}\\
truth        &    NM    &    EM    &   GG   &   FB   &    MN    &    CD   \\
\hline
NM           &  \textbf{0.68}    &  0.18    &  0.10  & 0.01   &  0.03    &  0.01   \\
EM           &  0.03    &  \textbf{0.91}    &  0.00  & 0.02   &  0.03    &  0.01   \\
GG           &  0.06    &  0.01    &  \textbf{0.70}  & 0.09   &  0.06    &  0.08   \\
FB           &  0.01    &  0.02    &  0.05  & \textbf{0.83}   &  0.05    &  0.05   \\
MN           &  0.09    &  0.00    &  0.07  & 0.04   &  \textbf{0.79}    &  0.00   \\
CD           &  0.02    &  0.01    &  0.10  & 0.18   &  0.01    &  \textbf{0.68}   \\
\hline
\end{tabular}
}
\end{center}
\caption{Confusion matrix for ILD classification (patch-level) with five-fold CV using GoogLeNet-TL.}
\label{tab:ild_confusion_matrix}
\end{table}

We evaluate the ILD classification task with five-fold CV on patient-level split, as it is more informative for real clinical performance than LOO.
The classification accuracy rates for interstitial lung disease detection are shown in Table \ref{tab:ild_accuracies}.
Two sub-tasks on ILD patch and slice classifications are conducted.
In general, patch-level ILD classification is less challenging than slice-level classification, as far more data samples can be sampled from the manually annotated ROIs (up to 100 image patches per ROI), available from \cite{depeursinge2012building}.
From Table \ref{tab:ild_accuracies}, all five deep models evaluated obtain comparable results within the range of classification accuracy rates $ [0.74,0.76]$.
Their averaged model achieves a slightly better accuracy of 0.79.

F1-scores \cite{song2013feature,song2015large,li2014medical} and the confusion matrix (Table \ref{tab:ild_confusion_matrix}) for patch-level ILD classification using GoogLeNet-TL under five-fold cross-validation (we denote as Patch-CV5) are also computed.
F1-scores are reported on patch classification only ($32\times 32$ pixel patches extracted from manual ROIs) \cite{song2013feature,song2015large,li2014medical}, as shown in Table \ref{tab:ild_f1_scores}. Both \cite{song2013feature} and \cite{song2015large} use the evaluation protocol of ``leave-one-patient-out'' (LOO), which is arguably much easier and not directly comparable to 10-fold CV \cite{li2014medical} or our Patch-CV5.
In this study, we classify six ILD classes by adding a consolidation (CD) class to five classes of healthy (normal - NM), emphysema (EM), ground glass (GG), fibrosis (FB), and micronodules (MN) in \cite{song2013feature,song2015large,li2014medical}.
Patch-CV10 \cite{li2014medical} and Patch-CV5 report similar medium to high F-scores. This implies that the ILD dataset (although one of the mainstream public medical image datasets) may not adequately represent ILD disease CT lung imaging patterns, over a population of only 120 patients.
Patch-CV5 yields higher F-scores than  \cite{li2014medical} and classifies the extra consolidation (CD) class.
At present, the most pressing task is to drastically expand the dataset or to explore across-dataset deep learning on the combined ILD and LTRC datasets \cite{Holmes2006}.

Recently, Gao et al. \cite{gao2014holistic} have argued that a new CADe protocol on holistic classification of ILD diseases directly, using axial CT slice attenuation patterns and CNN, may be more realistic for clinical applications.
We refer to this as slice-level classification, as image patch sampling from manual ROIs can be completely avoided (hence, no manual ROI inputs will be provided).
The experimental results in \cite{gao2014holistic} are conducted with a patient-level hard split of 100 (training) and 20 (testing).
The method's testing F-scores  (i.e., Slice-Test) are given in Table \ref{tab:ild_f1_scores}.
Note that the F-scores in \cite{gao2014holistic} are not directly comparable to our results, due to different evaluation criteria. Only Slice-Test is evaluated and reported in \cite{gao2014holistic}, and we find that F-scores can change drastically from different rounds of the five-fold CV.

\begin{figure*}[t]
\begin{center}
   \includegraphics[width=1\linewidth]{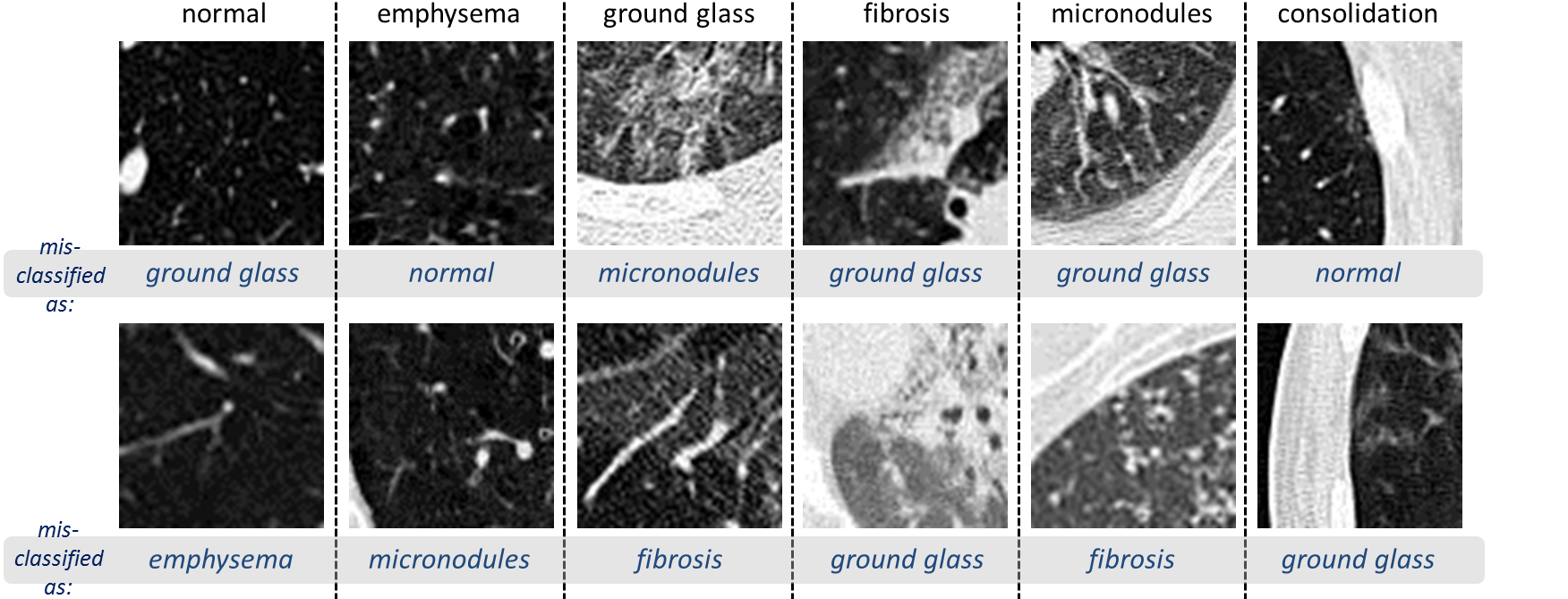}
\end{center}
   \caption{Visual examples of misclassified ILD 64x64 patches (in axial view), with their ground truth labels and inaccurately classified labels.}
\label{fig:miscls_exs_ilds}
\end{figure*}

While it is a more practical CADe scheme, slice-level CNN learning \cite{gao2014holistic} is very challenging, as it is restricted to only 905 CT image slices with tagged ILD labels. We only benchmark the slice-level ILD classification results in this section. Even with the help of data augmentation (described in Sec. \ref{sec:datasets}), the classification accuracy of GoogLeNet-TL from Table \ref{tab:ild_accuracies} is only 0.57. However, transfer learning from ImageNet pre-trained model is consistently beneficial, as evidenced by AlexNet-TL (0.46) versus AlexNet-RI (0.44), and GoogLeNet-TL (0.57) versus GoogLeNet-RI (0.41). It especially prevents GoogLeNet from over-fitting on the limited CADe datasets. Finally, when the cross-validation is conducted by randomly splitting the set of all 905 CT axial slices into five folds, markedly higher F-scores are obtained (Slice-Random in Table \ref{tab:ild_f1_scores}). This further validates the claim that the dataset poorly generalizes ILDs for different patients. Figure \ref{fig:miscls_exs_ilds} shows examples of misclassified ILD patches (in axial view), with their ground truth labels and inaccurately classified labels.

No existing work has reached the performance requirements for a realistic clinical setting \cite{gao2014holistic}, in which simple ROI-guided image patch extraction and classification (which requires manual ROI selection by clinicians) is implemented. The main goal of this paper is to investigate the three factors (CNN architectures, dataset characteristics and transfer learning) that affect performance on a specific medical image analysis problem and to ultimately deliver clinically relevant results. For ILD classification, the most critical performance bottlenecks are the challenge of cross-dataset learning and the limited patient population size. We attempt to overcome these obstacles by merging the ILD \cite{depeursinge2012building} and LTRC datasets.
Although the ILD \cite{depeursinge2012building} and LTRC datasets \cite{Holmes2006} (used in \cite{hofmanninger2015mapping}) were generated and annotated separately, they contain many common disease labels. For instance, the ILD disease classes emphysema (EM), ground glass (GG), fibrosis (FB), and micronodules (MN) belong  to both datasets, and thus can be jointly trained/tested to form a larger and unified dataset.

Adapting fully convolutional CNN or FCNN to parse every pixel location in the ILD lung CT images or slices, or adapting other methods from CNN based semantic image segmentation using PASCAL or ImageNet, may improve  accuracy and efficiency. However, current FCNN approaches \cite{Long_2015_CVPR,chen2014semantic} lack adequate spatial resolution in their directly output label space. A segmentation label propagation method was recently proposed \cite{gao2016isbi} to provide full pixel-wise labeling of the ILD data images.
In this work, we sample image patches from the slice using the ROIs for the ILD provided in the dataset, in order to be consistent with previous methods in patch-level \cite{song2013feature,song2015large,li2014medical} and slice-level classification \cite{gao2014holistic}.

\subsection{Evaluation of Five CNN Models using ILD Classification}

In this work, we mainly focus on AlexNet and GoogLeNet. AlexNet is the first notably successful CNN architecture on the ImageNet challenge and has rekindled significant research interests on CNN. GoogLeNet is the state-of-the-art deep model, which has outperformed other notable models, such as AlexNet, OverFeat, and VGGNet \cite{sermaneticlr14,simonyan2014very} in various computer vision benchmarks. Likewise, a reasonable assumption is that OverFeat and VGGNet may generate quantitative performance results ranked between AlexNet's and GoogLeNet's. For completeness, we include the Overfeat and VGGNet in the following evaluations, to bolster our hypothesis.

\paragraph{Overfeat}
OverFeat is described in \cite{sermaneticlr14} as an integrated framework for using CNN for classification, localization and detection.
Its architecture is similar to that of AlexNet, but contains far more parameters (e.g., 1024 convolution filters in both ``conv4'' and ``conv5'' layers compared to 384 and 256 convolution kernels in the ``conv4'' and ``conv5'' layers of AlexNet), and operates more densely (e.g., smaller kernel size of 2 in ``pool2'' layer ``pool5'' compared to the kernel size 3 in ``pool2'' and ``pool5'' of AlexNet) on the input image. Overfeat is the winning model of the ILSVRC 2013 in detection and classification tasks.

\paragraph{VGGNet}
The VGGNet architecture is introduced in \cite{simonyan2014very}, where it is designed to significantly increase the depth of the existing CNN architectures with 16 or 19 layers. Very small $3\times 3$ size convolutional filters are used in all convolution layers with a convolutional stride of size 1, in order to reduce the number of parameters in deeper networks. Since VGGNet is substantially deeper than the other CNN models, VGGNet is more susceptible to the vanishing gradient problem \cite{hochreiter1998vanishing,hinton2006fast,bengio1994learning}. Hence, the network may be more difficult to train. Training the network requires far more memory and computation time than AlexNet. We use the 16 layer variant as our default VGGNet model in our study.

%VGGNet was the runner-up model of the ILSVRC 2014 to GoogLeNet model \cite{szegedy2014going}, where both 16 layer (VGG-16) and 19 layer (VGG-19) variants achieved same top-1 classification accuracy, with 19 layer model achieving slightly higher top-5 accuracy.

The classification accuracy results for ILD slice and patch level classification of five CNN architectures (CifarNet, AlexNet, Overfeat, VGGNet and GoogLeNet) are shown in Table \ref{tab:cnn_arch_evals_01}. Based on the analysis in Sec. \ref{sec:ILD_detection_results}, transfer learning is only used for the slice level classification task. From Table \ref{tab:cnn_arch_evals_01}, quantitative classification accuracy rates increase as the CNN model becomes more complex (CifarNet, AlexNet, Overfeat, VGGNet and GoogLeNet, in ascending order), for both ILD slice and patch level classification problems. The reported results validate our assumption that  OverFeat's and VGGNet’s performance levels fall between AlexNet's and GoogLeNet‘s (this observation is consistent with the computer vision findings). CifarNet is designed for images with smaller dimensions ($32\times 32$ images), and thus is not catered to classification tasks involving  $256\times 256$ images.

To investigate the performance difference between five-fold cross-validation (CV) in Sec. \ref{sec:ILD_detection_results} and leave-one-patient-out (LOO) validation, this experiment is performed under the LOO protocol. By comparing results in Table \ref{tab:ild_accuracies} (CV-5) to those in Table \ref{tab:cnn_arch_evals_01} (LOO), one can see that LOO’s quantitative performances are remarkably better than CV-5's. For example, in ILD slice-level classification, the accuracy level drastically increases from 0.46 to 0.867 using AlexNet-TL, and from 0.57 to 0.902 for GoogLeNet-TL.

CNN training is implemented with the Caffe \cite{jia2013caffe} deep learning framework, using a NVidia K40 GPU on Ubuntu 14.04 Linux OS.
%To preserve the original ImageNet pre-trained CNN models for transfer learning on gray-scale medical images, the gray-scale images were replicated in R/G/B channels and fed into CNN models taking three channel images. 
All models are trained for up to 90 epochs with early stopping criteria, where a model snapshot with low validation loss is taken for the final model. Other hyper-parameters are fixed as follows: momentum: 0.9; weight decay: 0.0005; and a step learning rate schedule with base learning rate of 0.01, decreased by a factor of 10 every 30 epochs. The image batch size is set to 128, except for GoogLeNet's (64) and VGG-16's (32), which are the maximum batch sizes that can fit in the NVidia K40 GPU with 12GB of memory capacity. Table \ref{tab:cnn_arch_evals_02} illustrates the training time and memory requirements of the five CNN architectures on ILD patch-based classification up to 90 epochs.

\begin{table}[t]
\begin{center}
\resizebox{1\linewidth}{!}{
\begin{tabular}{|c|| c | c || c|}
\hline
       Method     &  ILD-Slice & Method & ILD-Patch \\
\hline\hline
CifarNet    &  -      & CifarNet & 0.799       \\
\hline
AlexNet-TL     &  0.867  & AlexNet-TL    & 0.865      \\
\hline
Overfeat-TL    &  0.877  &  Overfeat-TL  & 0.879       \\
\hline
VGG-16-TL      &  0.90   &  VGG-16-TL  & 0.893       \\
\hline
GoogLeNet-TL   &  0.902  & GoogLeNet-TL    & 0.911        \\
\hline
\end{tabular}
}
\end{center}
\caption{Classification results on ILD and LN detection with LOO.}
\label{tab:cnn_arch_evals_01}
\end{table}

\begin{table}[t]
\begin{center}
\resizebox{1\linewidth}{!}{
\begin{tabular}{|c|| c | c | c | c | c |}
\hline
                   & CifarNet & AlexNet & Overfeat & VGG-16 & GoogLeNet \\
\hline\hline
Time      & 7m16s    & 1h2m    & 1h26m    & 20h24m & 2h49m     \\
\hline
Memory    & 2.25 GB   & 3.45 GB  & 4.22 GB   & 9.26 GB & 5.37 GB    \\
\hline
\end{tabular}
}
\end{center}
\caption{Training time and memory requirements of the five CNN architectures on ILD patch-based classification up to 90 epochs.}
\label{tab:cnn_arch_evals_02}
\end{table}

\subsection{Training with ``Equal Prior'' vs. ``Biased Prior''}

{Medical datasets are often ``biased'', in that the number of healthy samples is much larger than the number of diseased instances, or that the numbers of images per class are uneven. In ILD dataset, the number of fibrosis samples is about 3.5 times greater than the number of  emphysema samples.  The number of non-LNs is $3\sim 4$ times greater than the number of LNs in lymph node detection. Different sampling or resampling rates are routinely applied to  both ILD and LN detection to balance the data sample number or scale per class, as in\cite{roth2015improving}. We refer this as ``Equal Prior''. If we use the same sampling rate, that will lead to a ``Biased Prior'' across different classes.

Without loss of generality, after GoogLeNet is trained on the training sets under ``Equal'' or ``Biased'' priors, we compare its classification results on the balanced validation sets. Evaluating a classifier on a biased validation set will cause unfair assessment of its performance. For instance, a classifier that predicts every image patch as ``non-LN'' will still achieve a $70 \%$ accuracy rate on a biased set with 
$3.5$ times as many non-LN samples as LN samples. The classification accuracy results of GoogLeNet trained under two configurations are shown in Table \ref{tab:data_char_prior}. Overall, it achieves lower accuracy results when trained with a ``biased prior'' in both tasks, and the accuracy difference for ILD patch-based classification is small. %This indicates the current setting for the ILD patch-based detection is already good so that the classification result does not get affected much.

\begin{table}[h]
\begin{center}
\resizebox{.65\linewidth}{!}{
\begin{tabular}{| c || c | c | }
\hline
               & ILD-Slice &  ILD-Patch  \\
\hline\hline
Equal Prior   & 0.902        & 0.953  \\
\hline
Biased Prior  & 0.872        & 0.952  \\
\hline
\end{tabular}
}
\end{center}
\caption{Classification accuracies for ILD slice and LN patch-level detection with ``equal prior'' and ``biased prior'', using GoogLeNet-TL.}
\label{tab:data_char_prior}
\end{table}

\section{Analysis via CNN Learning Traces \& lulVisualization}

In this section, we determine and analyze, via CNN visualization, the reasons for which transfer learning is beneficial to achieve better performance on CAD applications.

\begin{figure*}[t]
\begin{center}
   \includegraphics[width=1\linewidth]{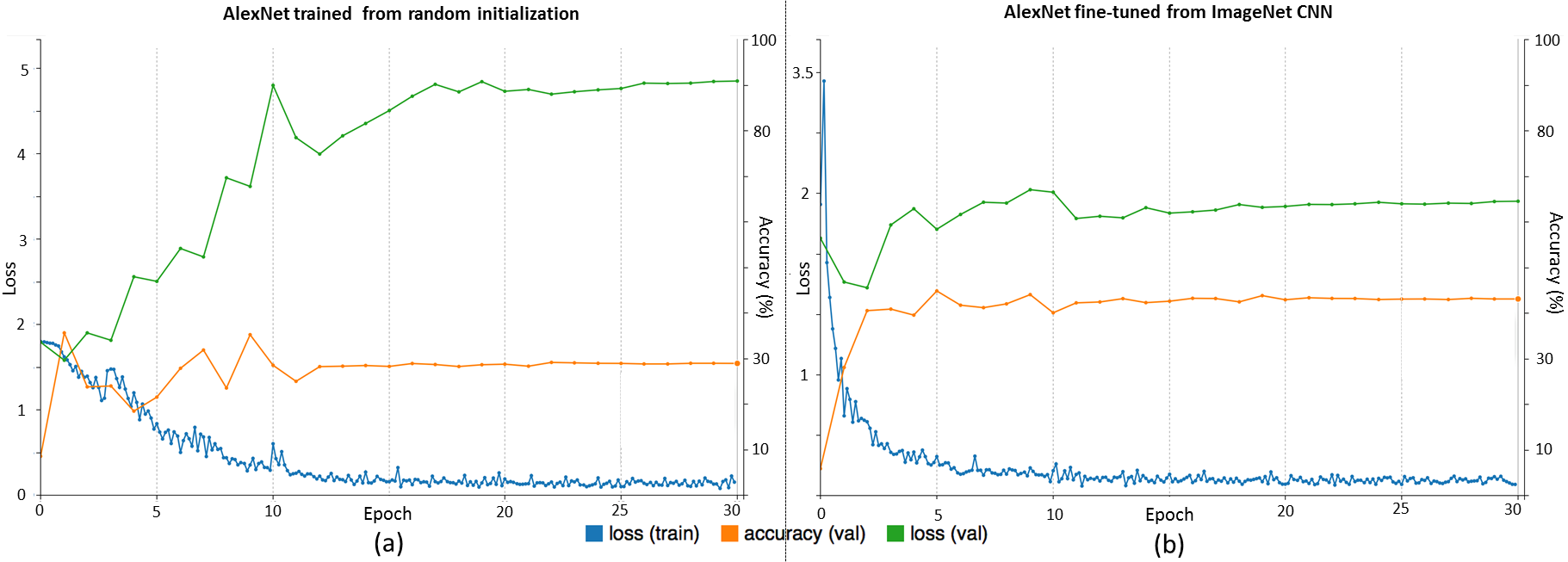}
\end{center}
   \caption{Traces of training and validation loss (blue and green lines) and validation accuracy (orange lines) during (a) training AlexNet from random initialization and (b) fine-tuning from ImageNet pre-trained CNN, for ILD classification.}
\label{fig:ild_alexnet_training_traces}
\end{figure*}

\textbf{Thoracoabdominal LN Detection.}
In Figure \ref{fig:ln_conv1s_RI}, the first layer convolution filters from five different CNN architectures are visualized.
We notice that without transfer learning \cite{razavian2014cnn,Girshick15}, somewhat blurry filters are learned (AlexNet-RI (256x256), AlexNet-RI (64x64), GoogLeNet-RI (256x256) and GoogLeNet-RI (64x64)).
However, in AlexNet-TL (256x256), many higher orders of contrast- or edge-preserving patterns (that enable capturing image appearance details) are evidently learned through fine-tuning from ImageNet.
With a smaller input resolution, AlexNet-RI (64x64) and GoogLeNet-RI (64x64) can learn image contrast filters to some degree; whereas, GoogLeNet-RI (256x256) and AlexNet-RI (256x256) have over-smooth low-level filters throughout.

\begin{figure*}[t]
\begin{center}
   \includegraphics[width=1\linewidth]{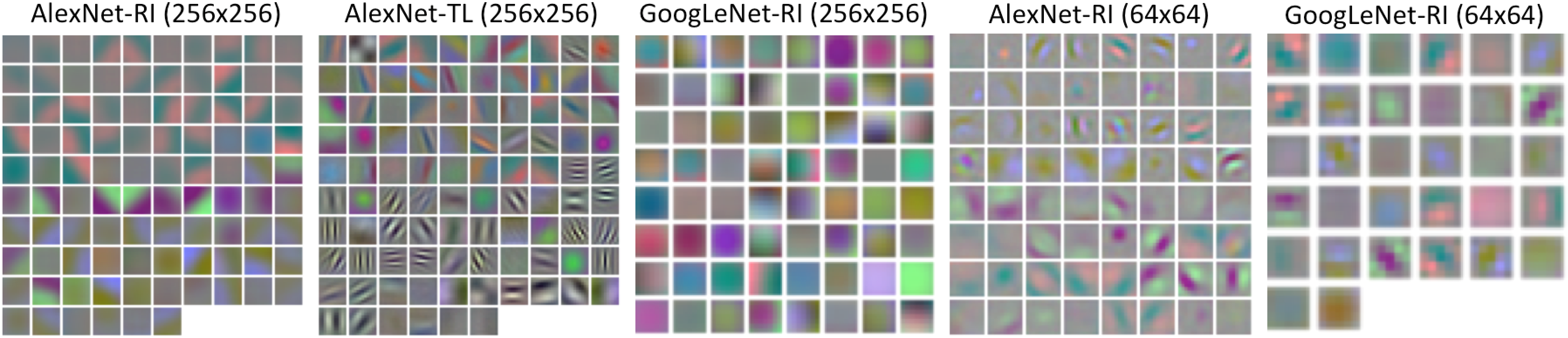}
\end{center}
   \caption{Visualization of first layer convolution filters of CNNs trained on abdominal and mediastinal LNs in RGB color, from random initialization (AlexNet-RI (256x256), AlexNet-RI (64x64), GoogLeNet-RI (256x256) and GoogLeNet-RI (64x64)) and with transfer learning (AlexNet-TL (256x256)).}
\label{fig:ln_conv1s_RI}
\end{figure*}

\textbf{ILD classification.} We focus on analyzing visual CNN optimization traces and activations from the ILD dataset, as its slice-level setting is most similar to ImageNet's. Indeed, both datasets use full-size images. The traces of the training loss, validation loss and validation accuracy of AlexNet-RI and AlexNet-TL, are shown in Figure \ref{fig:ild_alexnet_training_traces}. For AlexNet-RI in Figure \ref{fig:ild_alexnet_training_traces} (a), the training loss significantly decreases as the number of training epochs increases, while the validation loss notably increases and the validation accuracy does not improve much before reaching a plateau. With transfer learning and fine-tuning, much better and consistent performances of training loss, validation loss and validation accuracy traces are obtained (see Figure \ref{fig:ild_alexnet_training_traces} (b)). We begin the optimization problem -- that of fine-tuning the ImageNet pre-trained CNN to classify a comprehensive set of images -- by initializing the parameters close to an optimal solution. 
One could compare this process to making adults learn to classify ILDs, as opposed to babies.
During the process, the validation loss, having remained at lower values throughout, achieves higher final accuracy levels than the validation loss on a similar problem with random initialization. 
Meanwhile, the training losses in both cases decrease to values near zero. This indicates that both AlexNet-RI and AlexNet-TL over-fit on the ILD dataset, due to its small instance size. The quantitative results in Table \ref{tab:ild_accuracies} indicate that AlexNet-TL and GoogLeNet-TL have consistently better classification accuracies than AlexNet-RI and GoogLeNet-RI, respectively.

The last pooling layer (pool-5) activation maps of the ImageNet pre-trained AlexNet \cite{krizhevsky2012imagenet} (analogical to AlexNet-ImNet) and AlexNet-TL, obtained by processing two input images of Figure \ref{fig:ex_ilds} (b,c), are shown in Figure \ref{fig:viz_ild_ex} (a,b).
The last pooling layer activation map summarizes the entire input image by highlighting which relative locations or neural reception fields relative to the image are activated. There are a total of 256 (6x6) reception fields in AlexNet \cite{krizhevsky2012imagenet}.
Pooling units where the relative image location of the disease region is present in the image are highlighted with green boxes.
Next, we reconstruct the original ILD images using the process of de-convolution, back-propagating with convolution and un-pooling from the activation maps of the chosen pooling units \cite{zeiler2014visualizing}. From the reconstructed images (Figure \ref{fig:viz_ild_ex} bottom), we observe that with fine-tuning, AlexNet-TL detects and localizes objects of interest (ILD disease regions depicted in in Figure \ref{fig:ex_ilds} (b) and (c)) better than AlexNet-ImNet. The filters shown in Figure \ref{fig:viz_ild_ex} that better localize regions on the input images (Figure 2 (b) and (c)) respectively, produce relatively higher activations (in the top 5\%) among all 512 reception field responses in the fine-tuned AlexNet-TL model. As observed in \cite{Agrawal2015Analyzing}, the final CNN classification score can not be driven solely by a single strong activation in the receptions fields, but often by a sparse set of high activations (i.e., varying selective or sparse activations per input image).

\begin{figure*}[t]
\begin{center}
   \includegraphics[width=1\linewidth]{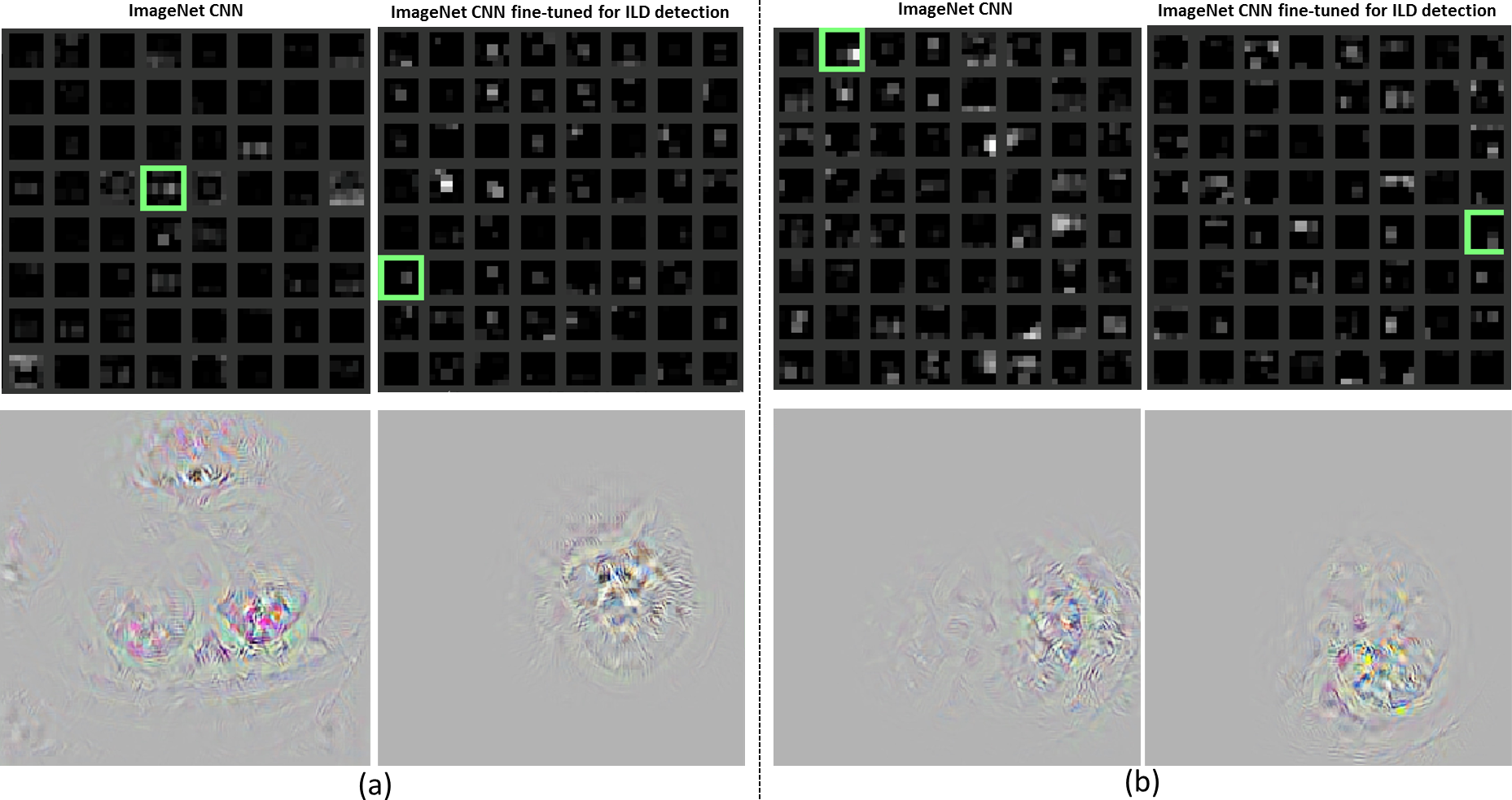}
\end{center}
   \caption{Visualization of the last pooling layer (pool-5) activations (top). Pooling units where the relative image location of the disease region is located in the image are highlighted with green boxes. The original images reconstructed from the units are shown in the bottom \cite{zeiler2014visualizing}. The examples in (a) and (b) are computed from the input ILD images in Figure \ref{fig:ex_ilds} (b) and (c), respectively.
   }
\label{fig:viz_ild_ex}
\end{figure*}

\section{Findings and Future Directions}

We summarize our findings as follows. 

\begin{itemize}
\item Deep CNN architectures with 8, even 22 layers \cite{krizhevsky2012imagenet,szegedy2014going}, can be useful even for CADe problems where the available training datasets are limited. Previously, CNN models used in medical image analysis applications have often been $2\sim5$ orders of magnitude smaller. 

\item The trade-off between using better learning models and using more training data \cite{Zhu2012do} should be carefully considered when searching for an optimal solution to any CADe problem (e.g., mediastinal and abdominal LN detection).
\item Limited datasets can be a bottleneck to further advancement of CADe. Building progressively growing (in scale), well annotated datasets is at least as crucial as developing new algorithms. This has been accomplished, for instance, in the field of computer vision. The well-known scene recognition problem has made tremendous progress, thanks to the steady and continuous development of Scene-15, MIT Indoor-67, SUN-397 and Place datasets \cite{zhou2014learning}. 

\item Transfer learning from the large scale annotated natural image datasets (ImageNet) to CADe problems has been consistently beneficial in our experiments. This sheds some light on cross-dataset CNN learning in the medical image domain, e.g., the union of the ILD \cite{depeursinge2012building} and LTRC datasets \cite{Holmes2006}, as suggested in this paper.

\item Finally, applications of off-the-shelf deep CNN image features to CADe problems can be improved by either exploring the performance-complementary properties of hand-crafted features \cite{Bar2015,Ginneken15,Ciompi2015automatic}, or by training CNNs from scratch and better fine-tuning CNNs on the target medical image dataset, as evaluated in this paper.
\end{itemize}

\section{Conclusion}

In this paper, we exploit and extensively evaluate three important, previously under-studied factors on deep convolutional neural networks (CNN) architecture, dataset characteristics, and transfer learning.
We evaluate CNN performance on two different computer-aided diagnosis applications: thoraco-abdominal lymph node detection and interstitial lung disease classification.
The empirical evaluation, CNN model visualization, CNN performance analysis, and conclusive insights can be generalized to the design of high performance CAD systems for other medical imaging tasks.

\section*{Acknowledgment}

This work was supported in part by the Intramural Research Program of the National Institutes of Health Clinical Center, and in part by a grant from the KRIBB Research Initiative Program (Korean Biomedical Scientist Fellowship Program), Korea Research Institute of Bioscience and Biotechnology, Republic of Korea.
This study utilized the high-performance computational capabilities of the Biowulf Linux cluster at the National Institutes of Health, Bethesda, MD (http://biowulf.nih.gov).
We thank NVIDIA for the K40 GPU donation.

% \end{thebibliography}
\bibliographystyle{IEEEtran}
%\fontsize{8}{9}\selectfont
\bibliography{tmi2015shin}

\begin{thebibliography}{10}
\providecommand{\url}[1]{#1}
\csname url@samestyle\endcsname
\providecommand{\newblock}{\relax}
\providecommand{\bibinfo}[2]{#2}
\providecommand{\BIBentrySTDinterwordspacing}{\spaceskip=0pt\relax}
\providecommand{\BIBentryALTinterwordstretchfactor}{4}
\providecommand{\BIBentryALTinterwordspacing}{\spaceskip=\fontdimen2\font plus
\BIBentryALTinterwordstretchfactor\fontdimen3\font minus
  \fontdimen4\font\relax}
\providecommand{\BIBforeignlanguage}[2]{{%
\expandafter\ifx\csname l@#1\endcsname\relax
\typeout{** WARNING: IEEEtran.bst: No hyphenation pattern has been}%
\typeout{** loaded for the language `#1'. Using the pattern for}%
\typeout{** the default language instead.}%
\else
\language=\csname l@#1\endcsname
\fi
#2}}
\providecommand{\BIBdecl}{\relax}
\BIBdecl

\bibitem{deng2009imagenet}
J.~Deng, W.~Dong, R.~Socher, L.-J. Li, K.~Li, and L.~Fei-Fei, ``Imagenet: A
  large-scale hierarchical image database,'' in \emph{IEEE CVPR}, 2009.

\bibitem{Russakovsky2014ILSVRC}
O.~Russakovsky, J.~Deng, H.~Su, J.~Krause, S.~Satheesh, S.~Ma, Z.~Huang,
  A.~Karpathy, A.~Khosla, M.~Bernstein, A.~Berg, and L.~Fei-Fei, ``Imagenet
  large scale visual recognition challenge,'' \emph{arXiv:1409.0575}, 2014.

\bibitem{LeCun1998gradient}
Y.~LeCun, L.~Bottou, Y.~Bengio, and P.~Haffner, ``Gradient-based learning
  applied to document recognition,'' \emph{Proc. of the IEEE}, vol.~86, no.~11,
  pp. 2278--2324, 1998.

\bibitem{krizhevsky2012imagenet}
A.~Krizhevsky, I.~Sutskever, and G.~E. Hinton, ``Imagenet classification with
  deep convolutional neural networks,'' in \emph{NIPS}, 2012.

\bibitem{krizhevsky2009learning}
A.~Krizhevsky, ``Learning multiple layers of features from tiny images,'' in
  \emph{Master's Thesis}.\hskip 1em plus 0.5em minus 0.4em\relax Dept. of Comp.
  Science, University of Toronto, 2009.

\bibitem{Girshick15}
R.~Girshick, J.~Donahue, T.~Darrell, and J.~Malik, ``Region-based convolutional
  networks for accurate object detection and semantic segmentation,''
  \emph{IEEE Trans. Pattern Anal. Mach. Intell.}, 2015.

\bibitem{He2015SPPNet}
K.~He, X.~Zhang, S.~Ren, and J.~Sun, ``Spatial pyramid pooling in deep
  convolutional networks for visual recognition,'' \emph{IEEE Trans. Pattern
  Anal. Mach. Intell.}, 2015.

\bibitem{Everingham2015Pascal}
M.~Everingham, S.~M.~A. Eslami, L.~Van~Gool, C.~Williams, J.~Winn, and
  A.~Zisserman, ``The pascal visual object classes challenge: A
  retrospective,'' \emph{International journal of computer vision}, vol. 111,
  no.~1, pp. 98--136, 2015.

\bibitem{Ginneken15}
B.~van Ginneken, A.~Setio, C.~Jacobs, and F.~Ciompi, ``Off-the-shelf
  convolutional neural network features for pulmonary nodule detection in
  computed tomography scans,'' in \emph{IEEE ISBI}, 2015, pp. 286--289.

\bibitem{Bar2015}
Y.~Bar, I.~Diamant, H.~Greenspan, and L.~Wolf, ``Chest pathology detection
  using deep learning with non-medical training,'' in \emph{IEEE ISBI}, 2015.

\bibitem{Shin2015Interleaved}
H.~Shin, L.~Lu, L.~Kim, A.~Seff, J.~Yao, and R.~Summers, ``Interleaved
  text/image deep mining on a large-scale radiology image database,'' in
  \emph{IEEE Conf. on CVPR}, 2015, pp. 1--10.

\bibitem{Ciompi2015automatic}
F.~Ciompi, B.~de~Hoop, S.~J. van Riel, K.~Chung, E.~Scholten, M.~Oudkerk,
  P.~de~Jong, M.~Prokop, and B.~van Ginneken, ``Automatic classification of
  pulmonary peri-fissural nodules in computed tomography using an ensemble of
  2d views and a convolutional neural network out-of-the-box,'' \emph{Medical
  Image Analysis}, 2015.

\bibitem{menze2014multimodal}
B.~Menze, M.~Reyes, and K.~Van~Leemput, ``The multimodal brain tumor image
  segmentation benchmark (brats),'' \emph{Medical Imaging, IEEE Trans. on},
  vol.~34, no.~10, pp. 1993--2024, 2015.

\bibitem{pan2015brain}
Y.~Pan, W.~Huang, Z.~Lin, W.~Zhu, J.~Zhou, J.~Wong, and Z.~Ding, ``Brain tumor
  grading based on neural networks and convolutional neural networks,'' in
  \emph{IEEE EMBC}, 2015, pp. 699--702.

\bibitem{shen2015multi}
W.~Shen, M.~Zhou, F.~Yang, C.~Yang, and J.~Tian, ``Multi-scale convolutional
  neural networks for lung nodule classification,'' in \emph{IPMI}, 2015, pp.
  588--599.

\bibitem{carneiro2015unregistered}
G.~Carneiro, J.~Nascimento, and A.~P. Bradley, ``Unregistered multiview
  mammogram analysis with pre-trained deep learning models,'' in \emph{MICCAI},
  2015, pp. 652--660.

\bibitem{wolterink2015automatic}
J.~M. Wolterink, T.~Leiner, M.~A. Viergever, and I.~I{\v{s}}gum, ``Automatic
  coronary calcium scoring in cardiac ct angiography using convolutional neural
  networks,'' in \emph{MICCAI}, 2015, pp. 589--596.

\bibitem{schlegl2014unsupervised}
T.~Schlegl, J.~Ofner, and G.~Langs, ``Unsupervised pre-training across image
  domains improves lung tissue classification,'' in \emph{Medical Computer
  Vision: Algorithms for Big Data}.\hskip 1em plus 0.5em minus 0.4em\relax
  Springer, 2014, pp. 82--93.

\bibitem{hofmanninger2015mapping}
J.~Hofmanninger and G.~Langs, ``Mapping visual features to semantic profiles
  for retrieval in medical imaging,'' in \emph{IEEE Conf. on CVPR}, 2015.

\bibitem{Carneiro2013}
G.~Carneiro and J.~Nascimento, ``Combining multiple dynamic models and deep
  learning architectures for tracking the left ventricle endocardium in
  ultrasound data,'' \emph{IEEE Trans. Pattern Anal. Mach. Intell.}, vol.~35,
  no.~11, pp. 2592--2607, 2013.

\bibitem{Shen2014}
R.~Li, W.~Zhang, H.~Suk, L.~Wang, J.~Li, D.~Shen, and S.~Ji, ``Deep learning
  based imaging data completion for improved brain disease diagnosis,'' in
  \emph{MICCAI}, 2014.

\bibitem{roth2015improving}
H.~Roth, L.~Lu, J.~Liu, J.~Yao, A.~Seff, K.~M. Cherry, E.~Turkbey, and
  R.~Summers, ``Improving computer-aided detection using convolutional neural
  networks and random view aggregation,'' in \emph{IEEE Trans. on Medical
  Imaging}, 2016.

\bibitem{barbu2012automatic}
A.~Barbu, M.~Suehling, X.~Xu, D.~Liu, S.~K. Zhou, and D.~Comaniciu, ``Automatic
  detection and segmentation of lymph nodes from ct data,'' \emph{Medical
  Imaging, IEEE Trans. on}, vol.~31, no.~2, pp. 240--250, 2012.

\bibitem{feulner2013lymph}
J.~Feulner, S.~K. Zhou, M.~Hammon, J.~Hornegger, and D.~Comaniciu, ``Lymph node
  detection and segmentation in chest ct data using discriminative learning and
  a spatial prior,'' \emph{Medical image analysis}, vol.~17, no.~2, pp.
  254--270, 2013.

\bibitem{feuerstein2012mediastinal}
M.~Feuerstein, B.~Glocker, T.~Kitasaka, Y.~Nakamura, S.~Iwano, and K.~Mori,
  ``Mediastinal atlas creation from 3-d chest computed tomography images:
  application to automated detection and station mapping of lymph nodes,''
  \emph{Medical image analysis}, vol.~16, no.~1, pp. 63--74, 2012.

\bibitem{lu2014computer}
L.~Lu, P.~Devarakota, S.~Vikal, D.~Wu, Y.~Zheng, and M.~Wolf, ``Computer aided
  diagnosis using multilevel image features on large-scale evaluation,'' in
  \emph{Medical Computer Vision. Large Data in Medical Imaging}.\hskip 1em plus
  0.5em minus 0.4em\relax Springer, 2014, pp. 161--174.

\bibitem{Lu2011}
L.~Lu, J.~Bi, M.~Wolf, and M.~Salganicoff, ``Effective 3d object detection and
  regression using probabilistic segmentation features in ct images,'' in
  \emph{IEEE CVPR}, 2011.

\bibitem{Lu2008}
L.~Lu, A.~Barbu, M.~Wolf, J.~Liang, M.~Salganicoff, and D.~Comaniciu,
  ``Accurate polyp segmentation for 3d ct colonography using multi-staged
  probabilistic binary learning and compositional model,'' in \emph{IEEE CVPR},
  2008.

\bibitem{tajbakhsh2015computer}
N.~Tajbakhsh, M.~B. Gotway, and J.~Liang, ``Computer-aided pulmonary embolism
  detection using a novel vessel-aligned multi-planar image representation and
  convolutional neural networks,'' in \emph{MICCAI}, 2015.

\bibitem{lowe2004distinctive}
D.~G. Lowe, ``Distinctive image features from scale-invariant keypoints,''
  \emph{International journal of computer vision}, vol.~60, no.~2, pp. 91--110,
  2004.

\bibitem{dalal2005histograms}
N.~Dalal and B.~Triggs, ``Histograms of oriented gradients for human
  detection,'' in \emph{IEEE CVPR}, vol.~1, 2005, pp. 886--893.

\bibitem{torralba2008small}
A.~Torralba, R.~Fergus, and Y.~Weiss, ``Small codes and large image databases
  for recognition,'' in \emph{IEEE CVPR}, 2008, pp. 1--8.

\bibitem{szegedy2014going}
C.~Szegedy, W.~Liu, Y.~Jia, P.~Sermanet, S.~Reed, D.~Anguelov, D.~Erhan, and
  A.~Rabinovich, ``Going deeper with convolutions,'' in \emph{IEEE Conf. on
  CVPR}, 2015.

\bibitem{Chatfield2014return}
K.~Chatfield, K.~Simonyan, A.~Vedaldi, and A.~Zisserman, ``Return of the devil
  in the details: Delving deep into convolutional nets,'' in \emph{BMVC}, 2014.

\bibitem{chatfield2011devil}
K.~Chatfield, V.~S. Lempitsky, A.~Vedaldi, and A.~Zisserman, ``The devil is in
  the details: an evaluation of recent feature encoding methods.'' in
  \emph{BMVC}, 2011.

\bibitem{seff20152d}
A.~Seff, L.~Lu, A.~Barbu, H.~Roth, H.-C. Shin, and R.~Summers, ``Leveraging
  mid-level semantic boundary cues for computer-aided lymph node detection,''
  in \emph{MICCAI}, 2015.

\bibitem{depeursinge2012building}
A.~Depeursinge, A.~Vargas, A.~Platon, A.~Geissbuhler, P.-A. Poletti, and
  H.~M{\"u}ller, ``Building a reference multimedia database for interstitial
  lung diseases,'' \emph{Computerized medical imaging and graphics}, vol.~36,
  no.~3, pp. 227--238, 2012.

\bibitem{song2013feature}
Y.~Song, W.~Cai, Y.~Zhou, and D.~D. Feng, ``Feature-based image patch
  approximation for lung tissue classification,'' \emph{Medical Imaging, IEEE
  Trans. on}, vol.~32, no.~4, pp. 797--808, 2013.

\bibitem{song2015large}
Y.~Song, W.~Cai, H.~Huang, Y.~Zhou, D.~Feng, Y.~Wang, M.~Fulham, and M.~Chen,
  ``Large margin local estimate with applications to medical image
  classification.'' \emph{IEEE Trans. on Medical Imaging}, 2015.

\bibitem{gao2014holistic}
M.~Gao, U.~Bagci, L.~Lu, A.~Wu, M.~Buty, H.-C. Shin, H.~Roth, Z.~G. Papadakis,
  A.~Depeursinge, R.~Summers, Z.~Xu, and J.~D. Mollura, ``Holistic
  classification of ct attenuation patterns for interstitial lung diseases via
  deep convolutional neural networks,'' in \emph{MICCAI first Workshop on Deep
  Learning in Medical Image Analysis}, 2015.

\bibitem{seff20142d}
A.~Seff, L.~Lu, K.~M. Cherry, H.~R. Roth, J.~Liu, S.~Wang, J.~Hoffman, E.~B.
  Turkbey, and R.~Summers, ``2d view aggregation for lymph node detection using
  a shallow hierarchy of linear classifiers,'' in \emph{MICCAI}, 2014, pp.
  544--552.

\bibitem{lu2011coarse}
L.~Lu, M.~Liu, X.~Ye, S.~Yu, and H.~Huang, ``Coarse-to-fine classification via
  parametric and nonparametric models for computer-aided diagnosis,'' in
  \emph{ACM Conf. on CIKM}, 2011, pp. 2509--2512.

\bibitem{Farabet2013}
C.~Farabet, C.~Couprie, L.~Najman, and Y.~LeCun, ``Learning hierarchical
  features for scene labeling,'' \emph{IEEE Trans. Pattern Anal. Mach.
  Intell.}, vol.~35, no.~8, pp. 1915--1929, 2013.

\bibitem{mostajabi2014feedforward}
M.~Mostajabi, P.~Yadollahpour, and G.~Shakhnarovich, ``Feedforward semantic
  segmentation with zoom-out features,'' \emph{arXiv preprint arXiv:1412.0774},
  2014.

\bibitem{roth20152DeepOrgan}
H.~Roth, L.~Lu, A.~Farag, H.-C. Shin, J.~Liu, E.~Turkbey, and R.~Summers,
  ``Deeporgan: Multi-level deep convolutional networks for automated pancreas
  segmentation,'' in \emph{MICCAI}, 2015.

\bibitem{liang2015human}
X.~Liang, C.~Xu, X.~Shen, J.~Yang, S.~Liu, J.~Tang, L.~Lin, and S.~Yan, ``Human
  parsing with contextualized convolutional neural network,'' in \emph{IEEE
  ICCV}, 2015, pp. 1386--1394.

\bibitem{gao2016isbi}
M.~Gao, Z.~Xu, L.~Lu, I.~Nogues, R.~Summers, and D.~Mollura, ``Segmentation
  label propagation using deep convolutional neural networks and dense
  conditional random field,'' in \emph{IEEE ISBI}, 2016.

\bibitem{wang2013multi}
H.~Wang, J.~W. Suh, S.~R. Das, J.~B. Pluta, C.~Craige, P.~Yushkevich
  \emph{et~al.}, ``Multi-atlas segmentation with joint label fusion,''
  \emph{IEEE Trans. Pattern Anal. Mach. Intell.}, vol.~35, no.~3, pp. 611--623,
  2013.

\bibitem{oquab2015object}
M.~Oquab, L.~Bottou, I.~Laptev, and J.~Sivic, ``Is object localization for
  free?--weakly-supervised learning with convolutional neural networks,'' in
  \emph{IEEE CVPR}, 2015, pp. 685--694.

\bibitem{oquab2014learning}
M.~Oquab, L.~Bottou, I.~Laptev, and S.~Josef, ``Learning and transferring
  mid-level image representations using convolutional neural networks,'' in
  \emph{IEEE CVPR}, 2015, pp. 1717--1724.

\bibitem{Zhu2012do}
X.~Zhu, C.~Vondrick, D.~Ramanan, and C.~Fowlkes, ``Do we need more training
  data or better models for object detection?'' in \emph{BMVC}, 2012.

\bibitem{Ciresan2013}
D.~Ciresan, A.~Giusti, L.~Gambardella, and J.~Schmidhuber, ``Mitosis detection
  in breast cancer histology images with deep neural networks,'' in
  \emph{MICCAI}, 2013.

\bibitem{zhang2015deep}
W.~Zhang, R.~Li, H.~Deng, L.~Wang, W.~Lin, S.~Ji, and D.~Shen, ``Deep
  convolutional neural networks for multi-modality isointense infant brain
  image segmentation,'' \emph{NeuroImage}, vol. 108, pp. 214--224, 2015.

\bibitem{li2014medical}
Q.~Li, W.~Cai, X.~Wang, Y.~Zhou, D.~D. Feng, and M.~Chen, ``Medical image
  classification with convolutional neural network,'' in \emph{IEEE ICARCV},
  2014, pp. 844--848.

\bibitem{miller1995wordnet}
G.~A. Miller, ``Wordnet: a lexical database for english,'' \emph{Communications
  of the ACM}, vol.~38, no.~11, pp. 39--41, 1995.

\bibitem{jia2013caffe}
Y.~Jia, E.~Shelhamer, J.~Donahue, S.~Karayev, J.~Long, R.~B. Girshick,
  S.~Guadarrama, and T.~Darrell, ``Caffe: Convolutional architecture for fast
  feature embedding.'' in \emph{ACM Multimedia}, vol.~2, 2014, p.~4.

\bibitem{razavian2014cnn}
A.~S. Razavian, H.~Azizpour, J.~Sullivan, and S.~Carlsson, ``Cnn features
  off-the-shelf: an astounding baseline for recognition,'' in \emph{IEEE
  CVPRW}, 2014, pp. 512--519.

\bibitem{zhou2014learning}
B.~Zhou, A.~Lapedriza, J.~Xiao, A.~Torralba, and A.~Oliva, ``Learning deep
  features for scene recognition using places database,'' in \emph{NIPS}, 2014,
  pp. 487--495.

\bibitem{Gupta2014Learning}
S.~Gupta, R.~Girshick, P.~Arbeláez, and J.~Malik, ``Learning rich features
  from rgb-d images for object detection and segmentation,'' in \emph{ECCV},
  2014, pp. 345--360.

\bibitem{Gupta2015Indoor}
S.~Gupta, P.~Arbeláez, R.~Girshick, and J.~Malik, ``Indoor scene understanding
  with rgb-d images: Bottom-up segmentation, object detection and semantic
  segmentation,'' \emph{International Journal of Computer Vision}, vol. 112,
  no.~2, pp. 133--149, 2015.

\bibitem{Gupta2013Natural}
A.~Gupta, M.~Ayhan, and A.~Maida, ``Natural image bases to represent
  neuroimaging data,'' in \emph{ICML}, 2013, pp. 987--994.

\bibitem{Chen2015Automatic}
H.~Chen, Q.~Dou, D.~Ni, J.~Cheng, J.~Qin, S.~Li, and P.~Heng, ``Automatic fetal
  ultrasound standard plane detection using knowledge transferred recurrent
  neural networks,'' in \emph{MICCAI}, 2015, pp. 507--514.

\bibitem{lkim2014rsna}
L.~Kim, H.~Roth, L.~Lu, S.~Wang, E.~Turkbey, and R.~Summers, ``Performance
  assessment of retroperitoneal lymph node computer-assisted detection using
  random forest and deep convolutional neural network learning algorithms in
  tandem,'' in \emph{the 102nd Annual Meeting of Radiological Society of North
  America}, 2014.

\bibitem{Holmes2006}
D.~Holmes~III, B.~Bartholmai, R.~Karwoski, V.~Zavaletta, and R.~Robb, ``The
  lung tissue research consortium: an extensive open database containing
  histological, clinical, and radiological data to study chronic lung
  disease,'' in \emph{2006 MICCAI Open Science Workshop}, 2006.

\bibitem{Long_2015_CVPR}
J.~Long, E.~Shelhamer, and T.~Darrell, ``Fully convolutional networks for
  semantic segmentation,'' in \emph{IEEE CVPR}, 2015.

\bibitem{chen2014semantic}
L.-C. Chen, G.~Papandreou, I.~Kokkinos, K.~Murphy, and A.~L. Yuille, ``Semantic
  image segmentation with deep convolutional nets and fully connected crfs,''
  \emph{ICLR}, 2015.

\bibitem{sermaneticlr14}
P.~Sermanet, D.~Eigen, X.~Zhang, M.~Mathieu, R.~Fergus, and Y.~LeCun,
  ``Overfeat: Integrated recognition, localization and detection using
  convolutional networks,'' in \emph{ICLR}, 2014.

\bibitem{simonyan2014very}
K.~Simonyan and A.~Zisserman, ``Very deep convolutional networks for
  large-scale image recognition,'' \emph{ICLR}, 2014.

\bibitem{hochreiter1998vanishing}
S.~Hochreiter, ``The vanishing gradient problem during learning recurrent
  neural nets and problem solutions,'' \emph{Int. J. of Uncertainty, Fuzziness
  and Knowledge-Based Systems}, vol.~6, no.~02, pp. 107--116, 1998.

\bibitem{hinton2006fast}
G.~E. Hinton, S.~Osindero, and Y.-W. Teh, ``A fast learning algorithm for deep
  belief nets,'' \emph{Neural computation}, vol.~18, no.~7, pp. 1527--1554,
  2006.

\bibitem{bengio1994learning}
Y.~Bengio, P.~Simard, and P.~Frasconi, ``Learning long-term dependencies with
  gradient descent is difficult,'' \emph{Neural Networks, IEEE Transactions
  on}, vol.~5, no.~2, pp. 157--166, 1994.

\bibitem{zeiler2014visualizing}
M.~D. Zeiler and R.~Fergus, ``Visualizing and understanding convolutional
  networks,'' in \emph{ECCV}, 2014, pp. 818--833.

\bibitem{Agrawal2015Analyzing}
P.~Agrawal, R.~Girshick, and J.~Malik, ``Analyzing the performance of
  multilayer neural networks for object recognition,'' in \emph{ECCV}, 2014.

\end{thebibliography}

\end{document}